\documentclass[letterpaper, 10 pt, conference]{ieeeconf}  % Comment this line out if you need a4paper

\IEEEoverridecommandlockouts                              % This command is only needed if 
\let\labelindent\relax

\usepackage{cancel}
\usepackage[inline]{enumitem}

\usepackage{graphicx}
\usepackage[font=small]{caption}
\usepackage{subcaption}
\usepackage{hyperref}
\usepackage{color}
\usepackage{gensymb}
\usepackage{amsmath}
\usepackage{amssymb}
\usepackage{dsfont}
\usepackage{cite}
\usepackage{adjustbox}
\usepackage{mathtools}
\usepackage{float}
\usepackage{flushend}
\usepackage{hhline}
\usepackage[aboveskip=1pt]{subcaption}
\usepackage[verbose]{placeins}

\newcommand{\bel}[0]{\ensuremath{\mathrm{Bel}}}
\DeclareMathOperator{\Tr}{Tr}

\usepackage[dvipsnames]{xcolor}
\usepackage{booktabs}
\usepackage{multirow}
\usepackage[normalem]{ulem}

\usepackage{tikz}
\usetikzlibrary{positioning}
\usetikzlibrary{bayesnet}
\usetikzlibrary{calc}
\usepackage{pgf}
\usepackage[dvipsnames]{xcolor}

\usepackage{cleveref}[2012/02/15]% v0.18.4; 
\crefformat{footnote}{#2\footnotemark[#1]#3}

\usepackage{stfloats}

\title{\LARGE \bf
ObVi-SLAM: Long-Term Object-Visual SLAM
}

\newcommand{\rulesep}{\unskip\ \vrule\ }

\newcommand{\AAA}[1]{ \noindent\textcolor{red}{Amanda: #1}}
\newcommand{\tc}[1]{\textcolor{Periwinkle}{Taijing: #1}}

\newcommand{\jb}[1]{}
\newcommand{\RA}[1]{}
\newcommand{\SM}[1]{}

\newcommand{\modified}[1]{#1}

\newcommand{\hide}[1]{}

\newlength{\twosubht}
\newsavebox{\twosubbox}

\author{Amanda Adkins \and Taijing Chen \and Joydeep Biswas% <-this % stops a space
\thanks{\modified{The authors}\hide{Amanda Adkins, Taijing Chen, and Joydeep Biswas} are with the Department of Computer Science, The University of Texas at Austin, Austin, TX. Email:
        {\tt\small \{aaadkins, taijing, joydeepb\}@cs.utexas.edu }}
\thanks{\modified{This work is partially supported by the National Science Foundation (GRFP DGE-2137420, CAREER-2046955) and Amazon Lab126. Any opinions, findings, and conclusions expressed in this material are those of the authors and do not necessarily reflect the views of the sponsors.}}
}

\begin{document}

\maketitle
\thispagestyle{empty}
\pagestyle{empty}

%%%%%%%%%%%%%%%%%%%%%%%%%%%%%%%%%%%%%%%%%%%%%%%%%%%%%%%%%%%%%%%%%%%%%%%%%%%%%%%%

\begin{abstract}

Robots responsible for tasks over long time scales must be able to localize consistently and scalably amid geometric, viewpoint, and appearance changes. 
% \AAA{Is there a smoother way to say this?}
% Existing SLAM approaches often use low-level features that are not robust to such changes and are high-dimensional, leading to scaling challenges over long-term deployments. In contrast, object detections are robust to these variations and are lower dimensional, but most object-based SLAM systems target short-term indoor deployments with high observability of objects\AAA{Is this the best way to separate other approaches from ours?}. \AAA{Do we need to allude to other long-term SLAM approaches here?}
% Existing visual SLAM approaches rely on low-level feature descriptors that are not robust to such environmental changes and result in large map sizes that scale poorly over long-term deployments. In contrast, object detections are robust to environmental variations and lead to more compact representations, but object-based SLAM systems require dense object observations for accuracy and suffer from poor short-term accuracy with sparse or erroneous object detections \AAAcritical{This seems hard to substantiate / a bit of an overstatement}.
Existing visual SLAM approaches rely on low-level feature descriptors that are not robust to such environmental changes and result in large map sizes that scale poorly over long-term deployments. In contrast, object detections are robust to environmental variations and lead to more compact representations, but most object-based SLAM systems target short-term indoor deployments with close objects.
% We posit that a SLAM system could provide both long-term robustness and short-term accuracy by leveraging both object detections and low-level visual features.
% We posit that a SLAM system could provide both long-term robustness and short-term accuracy by leveraging both object detections and low-level visual features, and by reasoning about 
In this paper, we introduce ObVi-SLAM to overcome these challenges by leveraging the best of both approaches.  ObVi-SLAM uses low-level visual features for high-quality short-term visual odometry; and to ensure global, long-term consistency, ObVi-SLAM builds an uncertainty-aware long-term map of persistent objects and updates it after every deployment.
By evaluating ObVi-SLAM on data from 16 deployment sessions spanning different weather and lighting conditions, we empirically show that ObVi-SLAM generates accurate localization estimates consistent over long-time scales in spite of varying \hide{lighting and }appearance conditions. 

\end{abstract}

%%%%%%%%%%%%%%%%%%%%%%%%%%%%%%%%%%%%%%%%%%%%%%%%%%%%%%%%%%%%%%%%%%%%%%%%%%%%%%%%
\section{Introduction}

% \jb{The first two paragraphs can be combined and condensed to quickly cover the following points:
% 1. We are interested in long-term autonomous mapping and localization
% 2. Existing visual SLAM approaches are susceptible to failures from...
% 3. V-SLAM maps are also large, and scale poorly across multiple deployments}
% Mobile service robots need to be able to operate autonomously over long time periods\hide{periods of time}. SLAM systems used in long-term deployments must be able to cope with environmental changes that occur over such time scales, including geometric changes from moving and movable objects, appearance changes from lighting and seasonal variation, and viewpoint differences arising from different perspectives of an environment. In addition, long-term SLAM systems must be able to scale with time. 

% Most existing SLAM systems focus on short-term deployments, ignoring both scaling concerns and these environmental factors that could impact robustness. Many existing visual SLAM approaches use low-level visual features, like ORB detections, which are not robust to long-term environmental changes and are also high-dimensional, contributing to scaling challenges. Some approaches \AAA{TODO references} fix the map after an initial mapping deployment, allowing for scalability, but making them brittle to initial errors and environmental changes. Some SLAM approaches aim to learn features invariant to appearance and viewpoint changes, but such learning-based methods are highly dependent on having training data that matches the characteristics of the target environment. 

Mobile service robots need to be able to operate autonomously over long time periods\hide{periods of time}. Existing visual SLAM approaches are susceptible to failures caused by environmental changes that occur over such time scales, including geometric changes from moving and movable objects, appearance changes from lighting and seasonal variation, and viewpoint differences arising from different perspectives of an environment. Further, visual SLAM systems generally use maps composed of visual feature points, which results in large maps that do not scale well with time. 

\begin{figure}[tbp]
\centering
\includegraphics[width=\columnwidth]{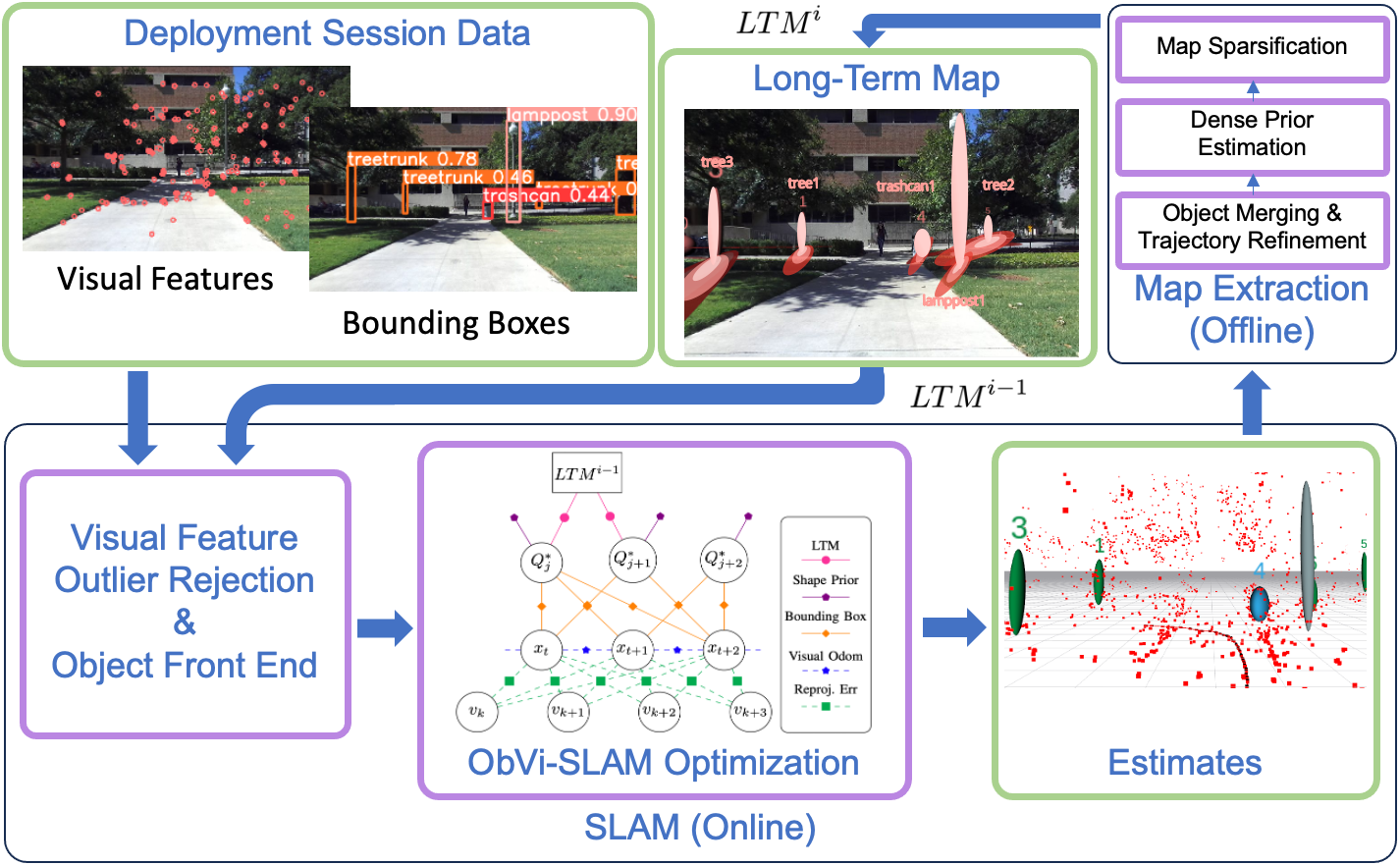}
\caption{ObVi-SLAM overview. In the factor graph, factors with solid lines are present for all optimizations, whereas use\hide{inclusion} of factors with dashed lines is dependent on if the optimization is local or global. }
\label{fig:approach_overview_graphical}
\vspace{-1em}
\end{figure}

% \jb{The ``more mature'' claim is hard to evaluate. Just mention that state-of-the-art object detectors are significantly more robust to viewpoint and lighting variations than feature matches with low-level visual feature descriptors.}
Unlike classical visual feature extractors, object detectors are trained to be invariant to appearance and large-scale viewpoint  changes\hide{ are more mature and available than learned models for SLAM systems}.
% \tc{``are trained be invariant to" seems a bit too strong to me personally. I'd hedge a bit if I'm writing it? Replace "invariant" with "robust" perhaps?}
Object-based SLAM systems \cite{cubeslam_Yang_2019, Nicholson2019QuadricSLAM,  ok2019ROSHAN, zins2022oa, soslamLiao2022 , scaleEstObjSong2022, eaoSLAM_wu_2020,  forwardObjInitialization2021Chen, outdoorInitandAssocTian2023, eqiJDACao2022} can leverage object detections to estimate landmarks that are similarly robust to such changes and provide a low-dimensional representation of the environment.  However, most existing object-based SLAM systems target short-term, indoor \hide{deployments and cover }trajectories in which objects can be viewed from diverse perspectives, simplifying \hide{the }object estimation\hide{problem}.

In this work, we introduce ObVi-SLAM: an object-visual SLAM approach designed for long-term deployments. ObVi-SLAM, shown in Fig. \ref{fig:approach_overview_graphical}, aims to generate accurate and consistent trajectory estimates for many deployment sessions over long time scales. To accomplish this objective, ObVi-SLAM utilizes a long-term map of the static objects in the target environment that reflects uncertainty of the current estimates. Use of object detectors trained to be invariant to different appearances enables ObVi-SLAM to be robust to appearance and viewpoint variation, while focusing on static objects provides resilience to movable entities. Forming the map with objects and their estimate uncertainty enables scalability over time and continuous improvement using new measurements from each deployment. Finally, use of low-level visual features tracked within individual deployment sessions enables high short-term accuracy. We summarize our contributions as follows:
\begin{enumerate*}[label=(\roman*)]
    \item a joint object-visual SLAM objective function for integrating long-term map information, object detections, and visual feature observations,
    \item a long-term scalable object front-end for initializing objects and associating incoming bounding boxes to existing objects, and
    \item a method for extracting a long-term map of the static objects in an environment with probabilistically-derived uncertainty estimates.
\end{enumerate*}
We empirically show that ObVi-SLAM can accurately and consistently estimate trajectories over long-term deployments using data collected outdoors across 16 deployment sessions outdoors over two months.

\section{Related Work}

\subsection{Object SLAM}
Initial efforts in semantic object SLAM such as SLAM++ \cite{salas2013slam++} used pre-existing models. Many later works either estimate detailed geometry for each instance \cite{grinvald2019volumetric, panotpicmultitsdf2022schmid} or use coarse, generalized models. For computational efficiency, ObVi-SLAM uses coarse models, thus we focus on the latter category.
% We focus on object SLAM approaches that use coarse generalized models for objects, rather than those that rely on pre-existing models or generate precise models at runtime \AAA{TODO references}. \AAA{TODO need any references to these? Leaning no since it doesn't really help reader understand our approach .} 
CubeSLAM \cite{cubeslam_Yang_2019} introduces an object-based SLAM algorithm that represents objects as cuboids and leverages relationships between low-level visual features, objects, and camera poses for estimation. 
% However, the CubeSLAM implementation relies on line detections originating from cuboidal object edges, limiting estimation of irregular objects. 
% \AAAcritical{Is it fair game to refer to the implementation here? The written paper notes a method for bounding-box only factors that aren't in the implementation}
Ellipsoids are also an increasingly popular representation for objects in SLAM. \hide{Several algorithms represent objects as ellipsoids. }QuadricSLAM \cite{Nicholson2019QuadricSLAM} is among the first to use ellipsoids to represent objects, but requires known data associations. 
% Since QuadricSLAM, numerous works have extended object SLAM using ellipsoids, focusing on aspects such as fast initialization and constraints to aid in cases of limited viewpoint diversity in ROSHAN \cite{ok2019ROSHAN}; trajectory reinitialization using objects in OA-SLAM \cite{zins2022oa}; object data association in EAO-SLAM \cite{eaoSLAM_wu_2020}; initialization for straight trajectory scenarios \cite{forwardObjInitialization2021Chen}; constraints for object arrangement, symmetry, and scale in SO-SLAM \cite{soslamLiao2022}; object initialization and association for outdoor SLAM \cite{outdoorInitandAssocTian2023}; and providing scale to monocular SLAM using object prior dimensions \cite{scaleEstObjSong2022}.
% Since QuadricSLAM, numerous works have 
Various extensions have enhanced object SLAM with ellipsoids, addressing aspects such as fast object initialization 
and constraints to aid in cases of limited viewpoint diversity 
in ROSHAN \cite{ok2019ROSHAN}; trajectory reinitialization using objects in OA-SLAM \cite{zins2022oa}; constraints for object arrangement, symmetry, and scale in SO-SLAM \cite{soslamLiao2022}; and  providing scale to monocular SLAM using object prior dimensions \cite{scaleEstObjSong2022}. Others focused on data association or initialization of ellipsoids, with ensemble data association \cite{eaoSLAM_wu_2020}; initialization for straight trajectory scenarios 
\cite{forwardObjInitialization2021Chen}; object initialization and association for outdoor SLAM \cite{outdoorInitandAssocTian2023}; and few-frame, limited viewpoint association and initialization \cite{eqiJDACao2022}.
% Ok et al. \cite{ok2019ROSHAN} build upon QuadricSLAM by introducing a fast initialization method and additional constrains to improve object estimation. OA-SLAM \cite{zins2022oa} estimates ellipsoids separately from the trajectory and introduces a method for initializing or relocalizing a camera from object detections and an object map. EAO-SLAM \cite{eaoSLAM_wu_2020} proposes an ellipsoid-based SLAM algorithm, with contributions focused on the data association problem. 
In contrast to the proposed approach, consideration of long-term SLAM is limited in these works: OA-SLAM is the only approach that addresses 
any aspects of long-term localization and does so with a fixed
map with no uncertainty estimation. 

\subsection{Long-Term Map Management}
SLAM approaches used for long-term robot deployments need to be able to scale over time. Many approaches have no support for multiple mapping sessions. Others \cite{ORBSLAM3_TRO, schneider2018maplab} support multi-session mapping, but do not address scaling concerns; another category fixes the map after initial data collection, addressing scaling\hide{ concerns} but leaving localization susceptible to initial errors and environmental changes. Some approaches, such as POCD \cite{QianChatrathPOCD}, maintain a fixed map but leverage change detection to determine when map regions should be updated. Other\hide{ SLAM} approaches prune or summarize the oldest data, retaining only the most informative or recent data.  Nashed and Biswas \cite{ltvm2016Nashed} generate an uncertainty-aware vector-based map of long-term features using lidar data from multiple deployments. A common technique for reducing map size used by ObVi-SLAM and numerous prior works is marginalization and sparsification.  Vallv\'e et al. \cite{factor_descent_marginalization_vallve_2018} proposed a method called factor descent for general SLAM marginalization and sparsification. Hsiung et al. \cite{info_theoretic_spars_hsiung_2018} and Wilbers et al. \cite{wilbers_global_sparse_prior_2019} both addressed the problem of online marginalization and sparsification for sliding window optimization; the former uses an information-theoretic approach applied to visual-inertial odometry, whereas the latter generates a single global prior to maintain optimal sparsity\hide{ while retaining some information from the marginalized nodes}. Zhao et al. \cite{lifelong_mapping_2021_zhao} focused  on long-term lidar\hide{-based} localization, using marginalization and sparsification to remove old submaps from a pose-graph. 

% Other relevant papers
% \cite{monoObjSlam2019Hosseinzadeh}

\subsection{Long-term Robust SLAM}

For a SLAM algorithm to be robust over long time scales, it must tolerate geometric, appearance, and viewpoint changes. 
Most approaches that aim to be robust to geometric changes arising from movable and moving entities aim to identify observations that may correspond to such objects, either using observed motion or semantic class.
Some of these approaches, such as DOT \cite{DOT_ballester_2021}, discard observations arising from these features, whereas others like CubeSLAM \cite{cubeslam_Yang_2019} and DynaSLAM II \cite{bescos2021dynaslam} use these objects for local corrections, estimating and tracking moving objects.
% Potential relevant citation \cite{drgslamWang2022} (use semantic segmentation + geometry to filter out dynamic entities)

% \AAA{TODO topic sentence for viewpoint and perspective change}.
Engineered feature representations, such as ORB features, can tolerate minor viewpoint and appearance changes. Recently, deep-learning has been applied to improve robustness to such changes. Gridseth et al. \cite{gridseth_2022_deep_learned_vtr} developed and trained a network to select sparse keypoints{, with a differentiable pose estimation framework}. Chen and Barfoot \cite{selfSupFeatsChen2023} similarly aimed to generate lighting-invariant keypoints in a self-supervised manner. Deja-Vu  \cite{spencer2020same} generates dense feature maps robust to seasonal changes using pairs of images captured in the same location in different conditions.  Learning-based methods like these, however, are highly dependent on having training data, often requiring labels or ground-truth, that matches the characteristics of the target environment.

\section{System Overview and Preliminaries}

% \AAA{Questions to figure out
% \begin{itemize}
%     \item Do we want to note the two phase optimization anywhere?
% \end{itemize}}

% 3/4 page allocated
\subsection{Approach Overview}
\hide{The goal of ObVi-SLAM is to generate accurate and consistent trajectory estimates over many deployment sessions over long-time scales. This requires ObVi-SLAM to scale over time and to be robust to changes that occur in long-term deployments, such as lighting changes, appearance variations, and movement of semi-static entities. To accomplish this objective, ObVi-SLAM uses a long-term map of the static objects in the target environment that reflects uncertainty of the current estimates. Using objects and uncertainty estimates in the map allows maps to scale over time, static objects are selected to enable robustness to moving entities, and use of object detectors trained to be invariant to different appearances enables ObVi-SLAM to be robust to appearance variation. }

There are 2 phases in ObVi-SLAM: SLAM and long-term map extraction. SLAM runs while the robot is deployed, estimating the robot's trajectory and the map of visual features and objects. Following this, estimates are refined and then  long-term map extraction uses the SLAM results\hide{data collected and estimated during deployment} to update the long-term map for use in subsequent sessions. The full ObVi-SLAM system  is illustrated in Fig. \ref{fig:approach_overview_graphical}.

% \begin{figure}[tbp]
% % \includegraphics[width=\columnwidth]{images/ObVi-SLAMSystemOverview_video_icon.png}
% % \includegraphics[width=\columnwidth]{images/OverviewFig.png}
% \includegraphics[width=\columnwidth]{images/ObviSLAMOverviewFigBiggerFont.png}
% \caption{ObVi-SLAM system diagram. Key contributions are in red.}
% \label{fig:approach_overview}
% \end{figure}

\subsection{Representation, Notation, and Assumptions}
We first define notation and representations for variables and note key assumptions. The following notation is used\hide{ throughout the paper}:
\begin{itemize}
    % \item $K$ - Camera intrinsic matrix. 
    \item $i \in [1, L]$ -  deployment session index. 
    \item $x_t^i \in SE(3)$ - pose at time{stamp} $t$ in session $i$; $t \in [1, T^i]$.
    \item ${x_t^{i}}{'} \in SE(3)$ - estimate for $x_t^i$ after the most recent optimization; $t \in [1, T^i]$.
    \item $v^i_k \in \mathbb{R}^3$ - $k$th visual feature\hide{ estimate} in session $i$; $k \in [1, V^i]$.
    \item $s_m^i \in \mathbb{R}^2$ - $m$th visual feature observation in session $i$: pixel at which feature was observed; $m \in [1, M^i]$.
    \item $B_n^i \in \mathbb{R}^4$ - $n$th object observation in session $i$ represented as a bounding box parameterized by the minimum and maximum x and y pixels; $n \in [1, N^i]$.
    \item $LTM^i$ - Long-term map summarizing information gathered through session $i$.
    \item $Q^{i*}_j$ - Object $j$ in session $i$, modeled as an ellipsoid; $j \in [1, J^i]$. 
    \item $c^i_j$ - Semantic class of object $Q^{i*}_j$.
\end{itemize}
Unless otherwise noted, variables correspond to deployment session $i$. Object notation follows conventions in \cite{ok2019ROSHAN} and \cite{ Nicholson2019QuadricSLAM}.
% \jb{What are the inputs and outputs of LBA, global alignment, and long-term mapping?}
% \AAA{inputs/outputs for the SLAM portion are defined very explicitly in IV.A. I'll make sure the same is true for the LTM}

There are two supported parameterizations of the ellipsoid: upright and full-DOF. In both cases, there are \hide{three parameters for position and three parameters for dimension}three position parameters and three dimension parameters. The upright parameterization represents orientation with a single parameter for  yaw\hide{, whereas}; the full-DOF parameterization uses three parameters \hide{to describe}for an $SO(3)$ orientation. \hide{Choice of parameterization}Parameterization choice depends on possible object configurations\hide{ of objects} in the target environment. We use the dual quadric form for each ellipsoid, which defines the ellipsoid using \hide{its}the relationship to its tangent planes. In the dual form, an \hide{individual }ellipsoid $Q^{*}_j$ is represented by a \hide{symmetric }4x4 matrix. For details on this representation and how the \hide{position, orientation, and dimensions}parameters are encoded in $Q^{*}_j$,  refer to \cite{Nicholson2019QuadricSLAM} and \cite{ok2019ROSHAN}.

The only sensor data required for ObVi-SLAM are camera images; ObVi-SLAM supports arbitrary camera configurations, including monocular and stereo, given that relative camera poses\hide{ between the cameras} are known. 
In our formulation, we assume that the robot's pose is combined with the appropriate relative camera pose when assessing an observation likelihood\hide{the likelihood of an observation}. \hide{In addition, w}
We also assume that the robot's pose\hide{ of the robot} at the beginning of each deployment session is known. 
Finally, the system should be configured to use object detections of the static classes for 
the target environment and 
approximate mean and variation for the dimensions of each semantic class should be given.

\section{Online SLAM Formulation}
% 1 3/4 page allocated
\subsection{SLAM Factor Graph}
During SLAM operation for deployment session $i$, we aim to find the optimal estimates for the visual features, $v_{1:V}$; objects, $Q^{*}_{1:J}$; and robot poses, $x_{1:T}$. Optimization inputs\hide{Inputs to the optimization} are the visual feature observations, $s_{1:M}$; object detections, $B_{1:N}$; prior mean and covariances for the dimensions of classes of interest; and, \hide{if it is not the first deployment}if it exists, the long-term map from a previous session, $LTM^{i-1}$.  We use two different \hide{modes of }optimization modes: local adjustment and global adjustment. Local adjustment optimizes the most recent poses and\hide{ corresponding} observations to generate high-quality visual odometry estimates\hide{, with the initial poses in the window set constant for stability}. For local adjustment, we use the following model for belief\hide{ over the variables of interest}: 
\begin{align}
    \bel(&x_{1:T}, v_{1:V}, Q^{*}_{1:J}) \propto \underbrace{ p(s_{1:M} | x_{1:T}, v_{1:V})}_{\substack{\text{Reprojection Error}}} 
    \label{eqn:local_adjustment} \\ \nonumber
    & \underbrace{ p(B_{1:N} | Q^*_{1:J}, x_{1:T})}_{\substack{\text{Bounding Box Error}}}   \underbrace{p(Q^*_{1:J} |c_{1:J})}_{\substack{\text{Semantic Shape Prior}}}    
     \underbrace{ p(Q^*_{1:J} | LTM^{i-1})}_{\substack{\text{Long-term Map Prior}}}.
\end{align}
Global adjustment aims to shift the estimates generated by the local optimization for improved global accuracy by using data from the entire trajectory. For computational efficiency, we use a simplified form of the belief, given by
\begin{align}
    % \bel(x, v,& Q^*) \propto  \underbrace{ p(x)}_{\substack{\text{Visual} \\ \text{Odometry Error}}} \hspace{-0.57em} \\ \nonumber
    %  &
    %   \underbrace{ p(s_O | Q^*, x)}_{\substack{\text{Bounding} \\ \text{Box Error}}} \underbrace{p(Q^* |c)}_{\substack{\text{Semantic} \\ \text{Shape Prior}}}    \underbrace{ p(Q^* | LTM^{i-1})}_{\substack{\text{Long-term} \\ \text{Map Prior}}} .
% 
    \bel(&x_{1:T}, v_{1:V}, Q^{*}_{1:J}) \propto \underbrace{ p(x_{1:T})}_{\substack{\text{Visual Odometry Error}}} \hspace{-0.57em} \label{eqn:global_adjustment}
     \\ \nonumber
    & \underbrace{ p(B_{1:N} | Q^*_{1:J}, x_{1:T})}_{\substack{\text{Bounding Box Error}}}   \underbrace{p(Q^*_{1:J} |c_{1:J})}_{\substack{\text{Semantic Shape Prior}}}    
     \underbrace{ p(Q^*_{1:J} | LTM^{i-1})}_{\substack{\text{Long-term Map Prior}}}. 
\end{align}

A factor graph reflecting these models is shown in Fig. \ref{fig:approach_overview_graphical}. Given the appropriate form of the belief, ObVi-SLAM then uses a nonlinear least squares optimizer to find the estimates that satisfy the following optimization objective:
\begin{align}
     % x^*, v^*, Q^{*\hide{^*}} &= {\underset{x, v, Q^*}{\mathrm{argmax}}\; \bel(x, v, Q^*)} \\ &\equiv {\underset{x, v, Q^*}{\mathrm{argmin}}\; \left[-\log \bel(x, v, Q^*)\right]}.
      x^*_{1:T}, &v^*_{1:V}, Q^{*\hide{^*}}_{1:J} = {\underset{x_{1:T}, v_{1:V}, Q^*_{1:J}}{\mathrm{argmax}}\; \bel(x_{1:T}, v_{1:V}, Q^*_{1:J})} \\ &\equiv {\underset{x_{1:T}, v_{1:V}, Q^*_{1:J}}{\mathrm{argmin}}\; \left[-\log \bel(x_{1:T}, v_{1:V}, Q^*_{1:J})\right]}.
\end{align}
% The factors in this optimization\AAA{should 'this optimization' be replaced with (\ref{eqn:local_adjustment}) and (\ref{eqn:global_adjustment})} are described further in the next subsections.
The factors in (\ref{eqn:local_adjustment}) and (\ref{eqn:global_adjustment}) are detailed\hide{ further} in the next subsections.

\subsubsection{Reprojection Factor}

The reprojection factor uses low-level visual features to generate a smooth trajectory estimate across local frames. Consistent with previous indirect visual SLAM approaches, ObVi-SLAM estimates the 3D locations of features detected in images by matching detections across temporally close frames. It adjusts the robot's trajectory and current visual feature position estimates to minimize the distance between the detected feature locations and the reprojections of the features onto the images. The 
reprojection factor term in our optimization objective is the 
sum of terms for each feature detection:
\begin{align}
    % -\log p(s_f | x, v) &= \sum_{{f_m} \in s_f} -\log p(f_m | x_{f_m}, v_{f_m}),  \\
    %  &= \frac{1}{2} \sum_{{f_m} \in s_f} || \hat{f}_{m} - f_m ||^2_{\Sigma_{f_m}}
    -\log p(s_{1:M} | &x_{1:T}, v_{1:V}) = \sum_{m =1}^{M} -\log p(s_m | x_{s_m}, v_{s_m})  \\
     &= \frac{1}{2} \sum_{m =1}^{M} || \Pi(x_{s_m}, v_{s_m}) - s_m ||^2_{\Sigma_{s_m}},
\end{align}
where $x_{s_m}$ and $v_{s_m}$ are the robot's pose and feature\hide{ estimate} for the detection, $\Pi(\cdot, \cdot)$ is the projection function giving the pixel location of a 3D point, and $\Sigma_{{s_m}}$ is the detection covariance.

\subsubsection{Visual Odometry Factor}

In the global trajectory optimization, to improve computation time and reduce the impact of local minima, we use visual odometry factors instead of reprojection error factors.
This term penalizes deviations from the result of the local optimization in the relative poses between pairs of subsequent poses. 
This has the form
\begin{align}
    % -\log p(x) &= \sum_{t=1}^{T} -\log p(x_t | x_{t-1}, x'_t, x'_{t-1}), \\
    %  &= \sum_{t=1}^{T}  \frac{1}{2}|| \left(\left(x_t \ominus x_{t-1}\right) \ominus \left(x'_t \ominus x'_{t-1} \right)\right) ||^2_{\Sigma_t}
    -\log p(x&_{1:T}) = \sum_{t=2}^{T} -\log p(x_t | x_{t-1}, x'_t, x'_{t-1}) \\
     &= \sum_{t=2}^{T}  \frac{1}{2}|| \left(\left(x_t \ominus x_{t-1}\right) \ominus \left(x'_t \ominus x'_{t-1} \right)\right) ||^2_{\Sigma_t},
\end{align}
where \hide{$x'_t$ is the estimate of the robot's pose at timestep $t$ from the most recent local optimization, $x_t$ is the current estimate for the pose, and  }$\Sigma_t$ is the covariance for the position deviation. $\Sigma_t$ is scaled with the
magnitude of $x'_t \ominus x'_{t-1}$.
% previously estimated translation and rotation change between the pair of poses. 

\subsubsection{Bounding Box Factor}

The bounding box factor aims to update object estimates such that edges of the detected bounding boxes for an object are tangent to the ellipse generated by projecting the ellipsoid onto the camera plane. This factor enables ObVi-SLAM to relate the current trajectory to the long-term map and update object estimates with new information. The bounding box factor  contains a term for each bounding box detection as follows:
\begin{align}
% -\log p(s_O | Q^*, x) &= \sum_{B_n \in s_O} -\log p(B_n | Q^*_{B_n}, x_{B_n}) \\ &= \frac{1}{2} \sum_{B_n \in s_O} || \hat{B}_n - {B_n} ||^2_{\Sigma_{B_n}},
-\log p(B_{1:N} | &Q^*_{1:J}, x_{1:T}) = \sum_{n=1}^N \hspace{-1pt}-\log p(B_n | Q^*_{B_n}, x_{B_n})  \nonumber \\ &= \frac{1}{2} \sum_{B_n \in s_O} || \hat{B}_n - {B_n} ||^2_{\Sigma_{B_n}},
\end{align}
where \hide{$B_n$ is the $n$th object detection, $Q^*_{B_n}$ is the associated object, $x_{B_n}$ is the pose at which $B_n$ was observed, }$\hat{B}_n$ is the bounding box resulting from projecting the ellipsoid onto the camera plane and $\Sigma_{B_n}$ gives the uncertainty of the object detection. To handle occlusions resulting from the image boundary, we inflate the entries of $\Sigma_{B_n}$ corresponding to the occluded edge when a detected bounding box lies near the image boundary.

% To find $\hat{B}_n$, we find the dual quadric representation with respect to the observing camera\hide{ that observed the object}, project the dual quadric to a dual conic, and find the horizontal and vertical tangent lines to the conic, which are arranged in a\hide{ 4 element} vector giving the minimum and maximum corners of the projected bounding box.\tc{(Very minor) If we need to save space, I don't think we need the first sentence here as the process is explained in details by the followings.}
To find $\hat{B}_n$, we project the ellipsoid onto the camera and find the horizontal and vertical tangent lines.
As noted\hide{ above}, the object's dual quadric representation \hide{of the object }in\hide{with respect to} the world frame is \hide{given by }a 4x4 matrix $Q^{*}_j$. The object's dual quadric representation with respect to the camera frame ${}^CQ^*_{j}$ is then given by 
\begin{align}
    {}^CQ^*_{j} = X Q^{*}_j X^T,
\end{align}
where $X$ is a homogeneous transformation giving the world
frame pose with respect to the camera, derived from $x_{B_n}$\hide{, obtained by taking the inverse of the product of the current robot's pose $x_{B_n}$ and the pose of the observing camera with respect to the robot}.
This is  projected to a dual conic section $G^*_j \in \mathbb{R}^{3x3}$ with 
\begin{align}
    G^*_j = K {}^CQ^*_{j} K^T.
\end{align}
All tangent lines $l$ to a dual conic must obey the property 
\begin{align}
    l^T G^* l = 0. \label{eqn:conic_tangent_prop}
\end{align}
We can find the horizontal $l_u$ and vertical $l_v$ tangent lines to the conic section, which take the forms $l_u = [1, 0, -u]$ and $l_v = [0, 1, -v]$. $u$ and $v$ are then obtained using (\ref{eqn:conic_tangent_prop}), giving 
\begin{align}
    % \hat{u}_{\min}, \hat{u}_{\max} &= \frac{G^*_{1, 3} \pm \sqrt{{G^*_{1, 3}}^2 - G^*_{1, 1} G^*_{3, 3}}}{G^*_{3, 3}} \\
    % \hat{v}_{\min}, \hat{v}_{\max} &= \frac{G^*_{2, 3} \pm \sqrt{{G^*_{2, 3}}^2 - G^*_{2, 2} G^*_{3, 3}}}{G^*_{3, 3}}. 
    \hat{u}_{\min}, \hat{u}_{\max} &= \left(G^*_{1, 3} \pm \sqrt{{G^*_{1, 3}}^2 - G^*_{1, 1} G^*_{3, 3}}\right) / {G^*_{3, 3}} \label{eqn:x_recovery} \\
    \hat{v}_{\min}, \hat{v}_{\max} &= \left(G^*_{2, 3} \pm \sqrt{{G^*_{2, 3}}^2 - G^*_{2, 2} G^*_{3, 3}}\right) /{G^*_{3, 3}}. \label{eqn:y_recovery} 
\end{align}
These are stacked to create vector $\hat{B}_n$. 
It should be noted that when either the x or y axes of the camera intersect an ellipsoid, the generated conic is not a proper ellipse\footnote{If both the x and y axes intersect the ellipsoid, the camera lies within the ellipsoid and the resulting conic is an imaginary ellipse; otherwise, the projection is a hyperbola, with one branch coming from behind the camera.  Future work could  improve handling of hyperbolic projections.}, resulting in imaginary bounding box values. In these cases, we set the error to a fixed value.

\subsubsection{Semantic Shape Prior Factor}
The semantic shape prior is important for constraining an object's dimensions in cases of limited viewpoint diversity where the object's depth and dimension along the camera ray are under-constrained. 
This factor penalizes deviations from the mean dimension of an object's semantic class. 
% \tc{I'm not sure if this is a good leading sentence here as readers unfamiliar with ROSHAN might have questions on the reasons behind this factor and the potential problems this factor can cause. (Specifically, in our case objects that share the same class are very likely to share the same dimension; For classes with larger variation, we increase the covariance, etc.) If I were writing it, I'd put the next sentence to the front and starting this paragraph by ``The semantic shape prior factor constraints dimensions of an object with limited viewpoint diversity", and then explain that it penalizes deviations before we start to present the math.} 
% The semantic shape prior factor penalizes deviations from the mean dimension of an object's semantic class. 
% This factor is important for constraining an object's dimensions in cases of limited viewpoint diversity where the object's depth and dimension along the camera ray are under-constrained. 
\hide{As with previous terms, t}The semantic shape prior\hide{ factor} is a sum of terms, each penalizing \hide{an individual}a single object, with the form
\begin{align}
% -\log p(Q^* |c) = \sum_{Q^{*}_j \in Q*} || m_{\text{shape}}(c_j) - d(Q^{*}_j) ||^2_{\Sigma_{c_j}},
-\log p(Q^*_{1:J} |c_{1:J}) = \sum_{j=1}^J || m_{\text{shape}}(c_j) - d(Q^{*}_j) ||^2_{\Sigma_{c_j}},
\end{align}
where $m_\text{shape}$ is a function giving the mean dimension for an object of the given semantic class, $d(\cdot)$ gives the dimensions of an object, and $\Sigma_{c_j}$ reflects the variance of the dimensions of an object of class $c_j$.
This term was introduced by Ok et al. \cite{ok2019ROSHAN} and additional details can be found in their paper. 

\subsubsection{Long-Term Map Prior Factor}
The long-term map prior factor incorporates information from previous sessions in our object estimation. This serves as a prior for objects that have been previously observed, so that their latest estimates appropriately balance previous knowledge and uncertainty with new observations. The ObVi-SLAM framework is generic to any factorization relating\hide{ previously observed} objects to each other and to the map frame. The generic form for this factor is
\begin{align}
     % p(Q^* | LTM^{i-1}) =  \hspace{-1.5em}\prod_{S_k \in {LTM^{i-1}}} \hspace{-1.5em} N(f_{S_k}(Q^*_{LTM^{i-1}}) | \mu_{S_k}, \Sigma_{S_k}), \label{eqn:generic_sparse_ltm}
    p(Q^*_{1:J} | LTM^{i-1}) =  \hspace{-1.5em}\prod_{S_k \in {LTM^{i-1}}} \hspace{-1.5em} N(f_{S_k}(Q^{*}_{1:J^{i-1}}) | \mu_{S_k}, \Sigma_{S_k}), \label{eqn:generic_sparse_ltm}
\end{align}
where $S_k$ is a component factor of the long-term map, $f_{S_k}(\cdot)$ is a function operating on a subset of the previously observed objects, and $\mu_{S_k}$ and $\Sigma_{S_k}$ are the mean and covariance for factor $S_k$. The current ObVi-SLAM formulation uses an independent normal assumption for each previously observed object, so the long-term map prior factor described generally in (\ref{eqn:generic_sparse_ltm}) has the form 
\begin{align}
    % -\log p(Q^* | LTM^{i-1}) = \hspace{-1.5em} \sum_{Q^{*}_j \in Q^*_{LTM^{i-1}}} \hspace{-1.5em}  \frac{1}{2}|| \text{vec}(Q^{*}_j) - \mu_{j} ||^2_{\Sigma_j},
    -\log p(Q^*_{1:J} | LTM^{i-1}) = \hspace{-2.5pt} \sum_{j=1}^{J^{i-1}} \hspace{-1pt} \frac{1}{2}|| \text{v}(Q^{*}_j) - \mu_{S_j} ||^2_{\Sigma_{S_j}},
\end{align}
where \hide{$Q^*_{LTM^{i-1}}$ is the subset of objects in the long-term map generated in the previous session, }$\text{v}(\cdot)$ is a function that generates a vector of\hide{with the minimal number of} parameters for object $Q^{*}_j$, and  $\mu_{S_j}$ and $\Sigma_{S_j}$ are the object estimate mean and uncertainty based on previous sessions.

\subsection{Front-End}
To use this optimization formulation, a front-end is needed to provide initial estimates and associations relating observations over time. Visual feature associations and initial estimates are obtained from the front-end of ORB-SLAM2 \cite{murORB2}, which generates feature tracks based on ORB descriptors. 

Our object-front end maintains a set of pending objects, adding them to the map after accumulating sufficient observations exceeding a minimum confidence. This guards against false positive object detections and prevents poorly-informed object estimates from influencing the trajectory estimates. Object initialization uses identity for orientation and the mean dimensions for the semantic class. The object's initial position is along the ray through the bounding box center and the depth is the distance at which the mean height yields the detected bounding box height. 

Associating incoming bounding boxes to past observations is a two-step process. We first perform local appearance-based matching using ORB features for the detected bounding boxes.  If there are no matches for a bounding box, a new pending object is created. ObVi-SLAM can thus generate high-quality local estimates for objects using the same objective function as the main problem, but optimizing only pending objects. After attaining a good local estimate, ObVi-SLAM performs geometric matching to determine if the pending object matches any older objects. If the objects' centers are close, the pending object's detections are merged into the older object and the pending object is removed. If there are no  existing objects that match the pending estimate, a new object is created. This method allows ObVi-SLAM's object front-end to scale with time, as only the features for the most recent bounding boxes are kept, avoiding long-term aggregation of object appearance data.

\section{Long-Term Map Generation}
% 1/2 page allocated

After SLAM for a deployment session, ObVi-SLAM refines results before creating a long-term map for subsequent sessions. A final optimization runs over the full trajectory, first using the model in (\ref{eqn:global_adjustment}), and then further refining with \hide{the model in }(\ref{eqn:local_adjustment}). This refinement is reserved for the post-processing stage due to computational complexity. 

ObVi-SLAM then checks if any of the objects should be merged into a single object, as imperfect data association can yield multiple objects for the same physical entity\hide{, particularly when the object is observed from different perspectives\AAA{Can we make it more clear that this means 2 different time chunks?}}. To address this, ObVi-SLAM searches for pairs of objects that have the same semantic class and are in close proximity, then merges the observations. Global refinement and object merging are repeated until no pairs of objects can be merged. 

Following SLAM execution and refinement, ObVi-SLAM summarizes the information obtained into a space-efficient long-term object map containing information only for relationships between objects and the global coordinate frame. Inputs to the long-term map extraction are the estimates $\{Q^{i*}_{1:J^i}, x^i_{1:T^i}, v^i_{1:V^i}\}$ and associations between observations and estimates. \hide{\AAA{put this in the intro}This allows the long-term map to be more space-efficient than visual feature-based maps or point clouds, as well as to retain only information likely to be invariant to geometric and appearance changes, as it focuses on consistently-detectable static objects.} Long-term map extraction has two phases: dense prior computation and map sparsification.

\subsection{Dense Prior Computation}

The dense prior is a marginal distribution over the objects, with visual features and robot poses marginalized out.\hide{ As we model the optimization problem with a normal distribution, the marginal distribution is a normal distribution with the only blocks corresponding to objects from the full distribution.}
As we model the full optimization problem with a normal distribution, the dense long-term map thus takes the form 
\begin{align}
% p(&Q^* | {LTM}^{i^D}) = N(Q^* | \mu_D, \Sigma_D) \\&=  N\left(\begin{bmatrix} 
    % Q_1 \\
    % \vdots \\
    % Q_{M_{i}}
    % \end{bmatrix}  \middle| \begin{bmatrix} 
    % \mu_1 \\
    % \vdots \\
    % \mu_{M_{i}}
    % \end{bmatrix} ,  \begin{bmatrix} 
    % \Sigma_{11} &  \hspace{-2.5em} \dots \hspace{-2.5em}  & \Sigma_{1M_{i}} \\
    % \vdots &  & \vdots \\
    % \Sigma_{1M_{i}}^T &  \hspace{-2.5em} \dots \hspace{-2.5em} & \Sigma_{M_{i}M_{i}} 
    % \end{bmatrix}\right),  \nonumber
    p(&Q^*_{1:J^i} | {LTM}^{i^D}) = N(Q^*
_{1:J^i} | \mu_D, \Sigma_D) \\&=  N\left(\begin{bmatrix} 
    Q_1 \\
    \vdots \\
    Q_{J_{i}}
    \end{bmatrix}  \middle| \begin{bmatrix} 
    \mu_1 \\
    \vdots \\
    \mu_{J_{i}}
    \end{bmatrix} ,  \begin{bmatrix} 
    \Sigma_{11} &  \hspace{-2.5em} \dots \hspace{-2.5em}  & \Sigma_{1J_{i}} \\
    \vdots &  & \vdots \\
    \Sigma_{1J_{i}}^T &  \hspace{-2.5em} \dots \hspace{-2.5em} & \Sigma_{J_{i}J_{i}} 
    \end{bmatrix}\right).  \nonumber
\end{align}
\hide{where $M_i$ is the number of objects at the end of session $i$.}

$\mu_D$ and $\Sigma_D$ are obtained with an optimization over the factors to summarize in the long-term map and the Markov blanket of the states to remove. In our case, we marginalize over the visual feature estimates, $v_{1:V^i}$, and robot pose estimates, $x_{1:T^i}$, obtained in session $i$. Shape priors are repeated across sessions, so these factors are omitted. Using properties of marginals for a normal distribution\hide{As the full optimization is represented as a normal distribution}\hide{ over all of the states estimated and the marginal distribution over a subset of multivariate normal random variables is simply a normal distribution composed of the elements corresponding to the retained variables}, \hide{the marginal using }$\mu_D$ and $\Sigma_D$ are formed from blocks corresponding to objects from the full mean and covariance. Thus, $\mu_D$ is given by
\begin{align}
% \mu_D &= \underset{Q^*}{\mathrm{argmax}}\; \underbrace{ p(s_f | x, v)}_{\substack{\text{Reprojection} \\ \text{Error}}} 
%      \underbrace{ p(s_O | Q^*, x)}_{\substack{\text{Bounding} \\ \text{Box Error}}} 
%      \underbrace{ p(Q^* | LTM^{i-1})}_{\substack{\text{Long-term Map} \\ \text{Prior for Ses. $i-1$}}}  \label{eqn:ltm_opt}
\mu_D =  \underset{Q^*_{1:J}}{\mathrm{argmax}}\; & \underbrace{ p(s_{1:M} | x_{1:T}, v_{1:V})}_{\substack{\text{Reprojection Error}}}  \label{eqn:ltm_opt} \\ 
&\underbrace{ p(B_{1:N} | Q^*_{1:N}, x_{1:T})}_{\substack{\text{Bounding Box Error}}}
  \underbrace{ p(Q^*_{1:J} | LTM^{i-1})}_{\substack{\text{Long-term Map Prior}}}.  \nonumber
\end{align}
For a least squares optimization problem given by $y^* = \hide{\underset{y}}{\mathrm{argmin}}\; || f(y) - z ||^2,$
% \begin{align}
    % y^* = \underset{y}{\mathrm{argmin}}\; || f(y) - z ||^2,
% \end{align}
where $z$ is a measurement and $f(y)$ is a predicted measurement given the state $y$, the covariance estimate $\Sigma_y$ is given by $\Sigma_y = (J^T(y^*) J(y^*))^{-1}$,
% \begin{align}
    % \Sigma_y = (J^T(y^*) J(y^*))^{-1},
% \end{align}
where $J(y^*)$ is the Jacobian of $f$ evaluated at $y^*$. Thus, $\Sigma_D$ is obtained using the Jacobian of the optimization expressed in (\ref{eqn:ltm_opt}), evaluated at $\mu_D$, taking the blocks of the covariance matrix that correspond to relationships between static objects.\hide{\footnote{Ceres solver\cite{Agarwal_Ceres_Solver_2022} provides methods for obtaining the cross-covariance for individual parameter blocks, thus bypassing the full covariance matrix calculation. We use this to find the required blocks of the covariance matrix.}}

\subsection{Map Sparsification}

The dense prior above would increase the computational complexity of optimization in a subsequent session, as modern optimization libraries rely on sparsity\hide{ in the optimization}\hide{ and the dense prior has no sparsity between the objects}. To address this, we aim to find a sparsified long-term map that approximates the dense prior but has fewer correlations between objects in the map. 
Sparsification first requires a design step to determine the factor topology. 
The generic form for the sparsified long-term map is shown in (\ref{eqn:generic_sparse_ltm}).\hide{ Using this representation, c} Choosing the factor topology can be seen as selecting the functions $f_{S_k}$ that comprise this long-term map representation. Once the factor topology is determined, the parameters $\mu_{S_k}$ and $\Sigma_{S_k}$ need to be found. To make the sparse long-term map prior as close as possible to the dense prior, $\mu_{S_k}$ and $\Sigma_{S_k}$ are identified by finding the values that minimize the KL divergence between the dense prior $N(Q^* | \mu_D, \Sigma_D)$, and the sparse long-term map prior in (\ref{eqn:generic_sparse_ltm}). After $\mu_{S_k}$ and $\Sigma_{S_k}$ are identified, the values for $\Sigma_{S_k}$ may suggest that a component factor of the sparse long-term map does not carry significant information, in which case minimally meaningful components $f_{S_k}$ can be removed from the topology and the parameter identification can be repeated, encouraging further sparsity in $LTM^i$. 

As noted above, the current version of ObVi-SLAM uses the sparse prior form with independently, normally-distributed ellipsoids. $LTM^i$ thus takes the form
\begin{align}
% p(Q^* | LTM^i) &=N(Q^* | \mu_S, \Sigma_S) \\&= \prod_{Q^{i*}_j \in Q^*} N(Q^{i*}_j| \mu_{S_j}, \Sigma_{S_j}).
p(Q^*_{1:J^i} | LTM^i) &=N(Q^*_{1:J^i} | \mu_S, \Sigma_S) \\&= \prod_{j=1}^{J^i} N(Q^{*}_j| \mu_{S_j}, \Sigma_{S_j}).
\end{align}
Using this topology, the KL divergence between the dense long-term map and sparse long-term map is
\begin{align}
% D_{KL}&(p(Q^* | LTM^{i^D}) || p(Q^* | LTM^{i})) = \frac{1}{2}\Bigl(\Tr(\Sigma_S^{-1}\Sigma_D) \nonumber \\
% & - d + (\mu_S - \mu_D)^T 
% \Sigma_{S}^{-1} (\mu_S - \mu_D) + \ln\frac{|\Sigma_S|}{|\Sigma_D|} \Bigr),  
&D_{KL}(p(Q^*_{1:J^i} | LTM^{i^D}) || p(Q^*_{1:J^i} | LTM^{i})) = \frac{1}{2}\Bigl(\ln\frac{|\Sigma_S|}{|\Sigma_D|} \nonumber \\
& - d + (\mu_S - \mu_D)^T  \Sigma_{S}^{-1} (\mu_S - \mu_D) + \Tr(\Sigma_S^{-1}\Sigma_D) \Bigr),  
\label{eqn:kl_divergence}
\end{align}
with $\Sigma_S$ constrained to be a block diagonal matrix with entries $\Sigma_{S_j}$ and $d$ being the dimension of the normal distribution. \hide{We differentiate to find the values of $\mu_S$ and $\Sigma_S$ that minimize the KL divergence. In doing so, we find that the optimal $\mu_S$ is equal to $\mu_D$ and that the optimal $\Sigma_S$ has diagonal blocks that match the corresponding diagonal blocks in $\Sigma_D$, with the remainder of $\Sigma_S$ composed of $\mathbf{0}$ entries, as is consistent with independently distributed ellipsoids.}
Differentiating gives the optimal $\mu_S$ equal to $\mu_D$ and the optimal $\Sigma_S$ as a block diagonal matrix with the corresponding entries from in $\Sigma_D$, as is consistent with independently distributed ellipsoids.
More specifically,  this gives $\mu_{S_j}$ as the maximum likelihood estimate for the $j$th ellipsoid for (\ref{eqn:ltm_opt}) and $\Sigma_{S_j}$ as the covariance representing the uncertainty after optimization for the $j$th ellipsoid estimate.

\section{Experimental Results} 

% INTRO

% Experimental questions:
% \begin{itemize}
%     \item How globally accurate are trajectory estimates generated ObVi-SLAM for individual trajectories?
%     \item How consistent are localization estimates from ObVi-SLAM with respect to a global reference frame under diverse lighting conditions and appearances?
%     \item How accurate are object estimates generated by ObVi-SLAM?
%     \item How space-efficient are maps generated by ObVi-SLAM?
%     \item \AAA{Do we need experimental question for ablations?} What is the impact of each component of the formulation for ObVi-SLAM on its performance?
%     % \item \AAA{Do we have time/space for perturbations of object detections?} How does the quality of object detections impact the performance of ObVi-SLAM?
% \end{itemize}

We aim to answer the following questions in our evaluation: \begin{enumerate*}[label=(\roman*)]
    \item How globally accurate are individual trajectory estimates generated by ObVi-SLAM\hide{ for individual trajectories}?
    \item How consistent are ObVi-SLAM localization estimates with respect to a global \hide{reference }frame under diverse\hide{ lighting conditions and} appearances?
    \item How accurate are object estimates generated by ObVi-SLAM?
    \item How space efficient are ObVi-SLAM maps\hide{ generated by ObVi-SLAM}?
    \item What is the impact of each component of the formulation for ObVi-SLAM on its performance?
\end{enumerate*}
Parameters and \hide{source }code for ObVi-SLAM and our evaluation pipeline can be found in our github repository.\footnote{\label{footnote:website}Code + website: \url{https://github.com/ut-amrl/ObVi-SLAM}; Code uses Ceres Solver \cite{Agarwal_Ceres_Solver_2022}.}

% Figures
% \begin{itemize}
%     \item Translational ATE plot(vs others)
%     \item Rotational ATE plot (vs others)
%     \item Translational consistency CDF (vs others)
%     \item Rotational consistency CDF (vs others)
%     \item Qualitative plots -- trajectory overlays (for all approaches? for just ours?) Include images from subset of trajectories to highlight lighting change, include bounding boxes (point to trajectory segment where this happened)
%     \item Object position  (vs others) (CDF vs avg?) (for all or at the end of last session) (avg for every trajectory)
%     \begin{itemize}
%         \item How do we handle duplicated objects/false positives? (for iou take union of space)
%         \item Distance based on closest
%         \item Iou based on union
%         \item Duplication ratio
%     \end{itemize}
%     \item Object IoU  (CDF vs avg?) (for all or at the end of last session?) (ignore trees)
%     \item Repeat quantitative tables for ablations (i.e. all but trajectories overlayed)
% \end{itemize}

\subsection{Experiment Setup and Implementation Details}
\begin{figure}[tbp]
  \vspace{1em}
  \centering

    \includegraphics[width=\linewidth]{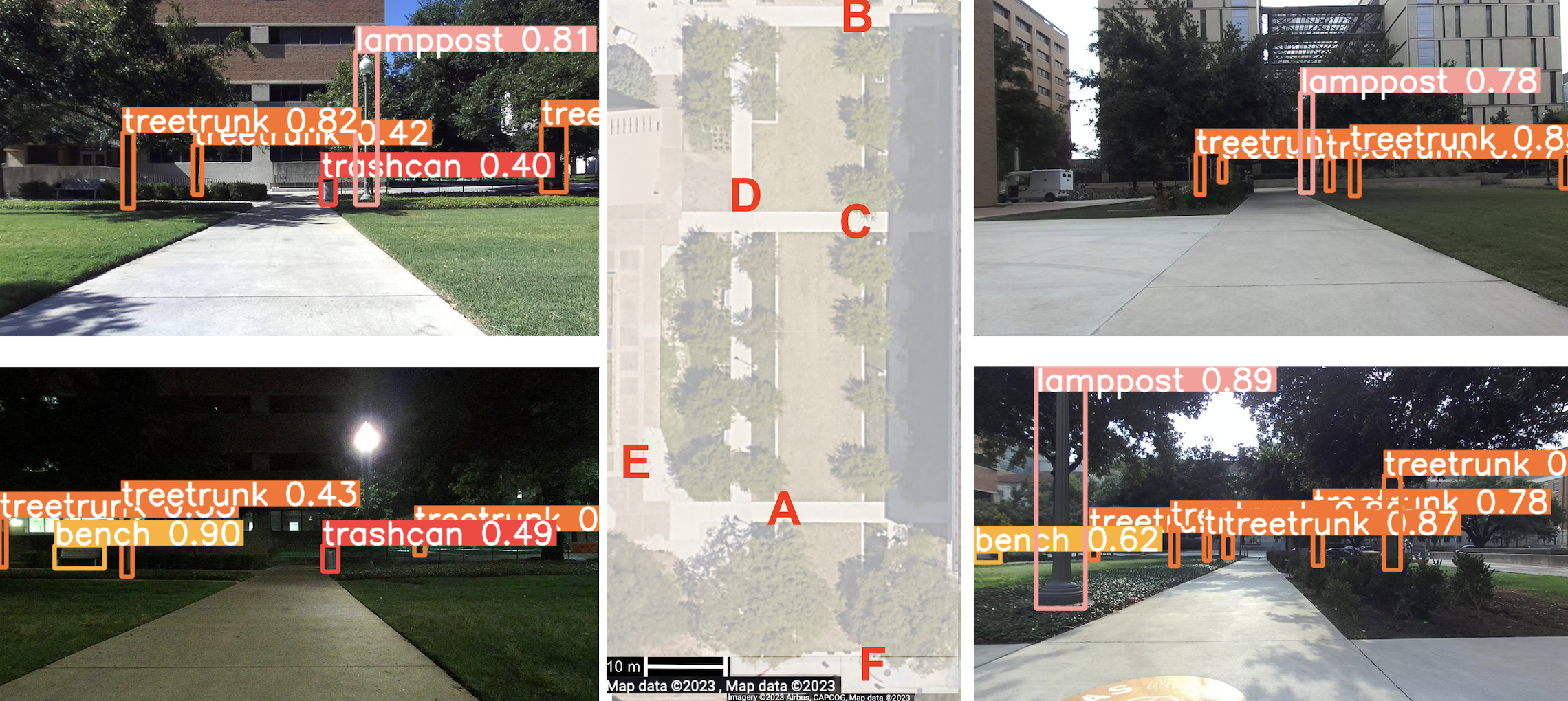}

  \caption{Test environment: appearance change at waypoint A; satellite view with waypoint labels; waypoint D from different perspectives.}
  \label{fig:test_env}
\end{figure}

To evaluate our approach\hide{ and others}, we collected 16 trajectories over two months in an outdoor campus environment, shown in Fig. \ref{fig:test_env},\hide{.  The trajectories were collected} with different lighting conditions, covering\hide{ and covered} \hide{a variety of}varying paths\hide{ through the environment} to capture different viewpoints. In each \hide{of the trajectories}trajectory, the robot started and ended at the same location, visiting a subset of six waypoints located consistently across the trajectories. 
% For object-based approaches, we
We
used YOLOv5 \cite{yolov5} with a custom model \hide{to provide}for object detections of tree trunks, lampposts, benches, and trash cans.

For our evaluations, we used the upright parameterization of objects for ObVi-SLAM. We compared ObVi-SLAM to ORB-SLAM3 \cite{ORBSLAM3_TRO}, a visual feature-based multi-session SLAM approach; OA-SLAM \cite{zins2022oa}, a joint object-visual SLAM approach that uses objects only for relocalization; and DROID-SLAM \cite{teed2021droid}, a learning-based SLAM approach. For all algorithms, we used stereo image data from a Zed2i mounted on a Clearpath Jackal. DROID-SLAM does not support map saving\hide{ and loading}, so each trajectory was evaluated individually\hide{ to test a high-performing mapless approach against the map-based approaches}. ORB-SLAM3 supports map saving, but maximum capacity for a map data structure was reached after the seventh trajectory, so the first seven trajectories were run in sequence and the remaining nine \hide{trajectories }were run individually starting with the map from  the seventh trajectory. This is denoted in Fig. \ref{fig:comps_ates}\hide{the per-trajectory results} with a dashed line. OA-SLAM was run with the same object detections as ObVi-SLAM and maps passed between sessions were aggregated across the 16 trajectories. Ground truth was obtained by running LeGO-LOAM \cite{legoloam2018} with an Ouster OS1 lidar and aligning 
the individual trajectories using the waypoints.

\subsection{Trajectory Accuracy}

\begin{figure}[tbp]
\centering
\begin{subfigure}[b]{0.49\columnwidth}
\centering
\includegraphics[width=\textwidth]{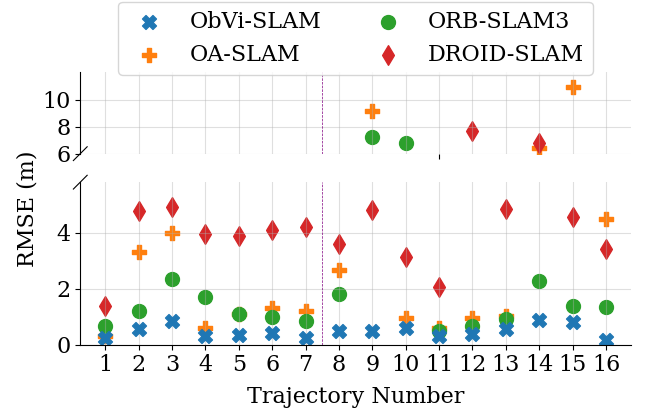}
\subcaption{Translation ATE.}
\label{fig:comps_transl_est_ate}
\end{subfigure}
\begin{subfigure}[b]{0.49\columnwidth}
\centering
\includegraphics[width=\textwidth]{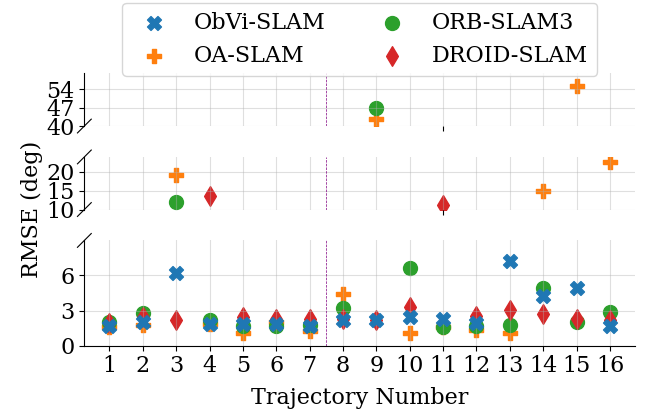}
\subcaption{Orientation ATE.}
\label{fig:comps_orient_est_ate}
\end{subfigure}
\caption{Translation and orientation ATE for ObVi-SLAM and comparison approaches for each trajectory.}
\label{fig:comps_ates}
\vspace{-2em}
\end{figure}

\begin{figure}[t!]
\vspace{1em}
\centering
\begin{subfigure}[b]{0.49\columnwidth}
\centering
\includegraphics[width=\textwidth]{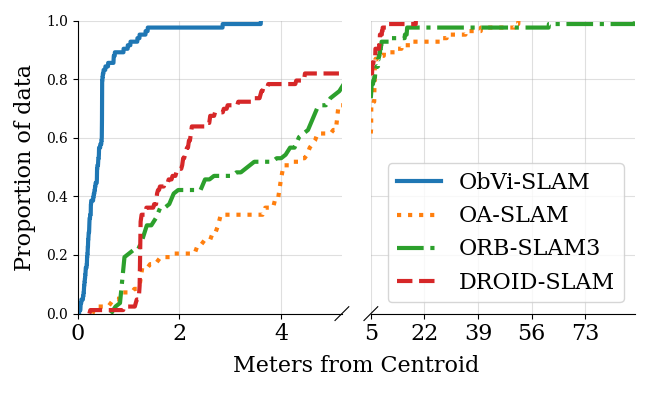}
\subcaption{CDF of position\hide{ estimate} deviation.}
\label{fig:comps_transl_est_cdf}
\end{subfigure}
\begin{subfigure}[b]{0.49\columnwidth}
\centering
\includegraphics[width=\textwidth]{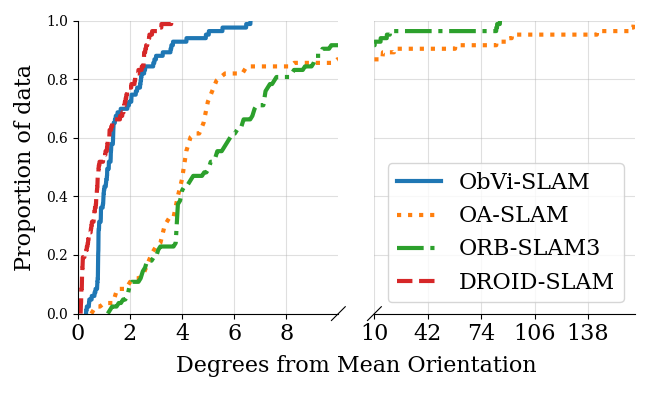}
\subcaption{CDF of orientation\hide{ estimate} deviation.}
\label{fig:comps_orient_est_cdf}
\end{subfigure}
\caption{Estimate\hide{Position and orientation} consistency for ObVi-SLAM and comparison approaches. An optimal algorithm would quickly rise to 1.}
\label{fig:comps_consist_cdfs}
\vspace{-2em}
\end{figure}

\begin{figure*}[t]
\vspace{1em}
% preliminary
\sbox\twosubbox{%
  \resizebox{\dimexpr.70\textwidth-1em}{!}{%
    \includegraphics[height=3cm]{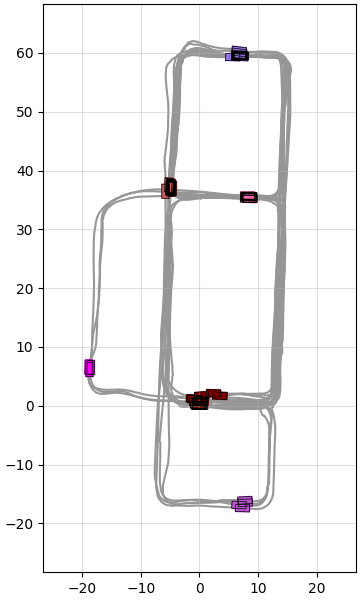}%
    \includegraphics[height=3cm]{images/TopdownTrajObVi.png}%
    \includegraphics[height=3cm]{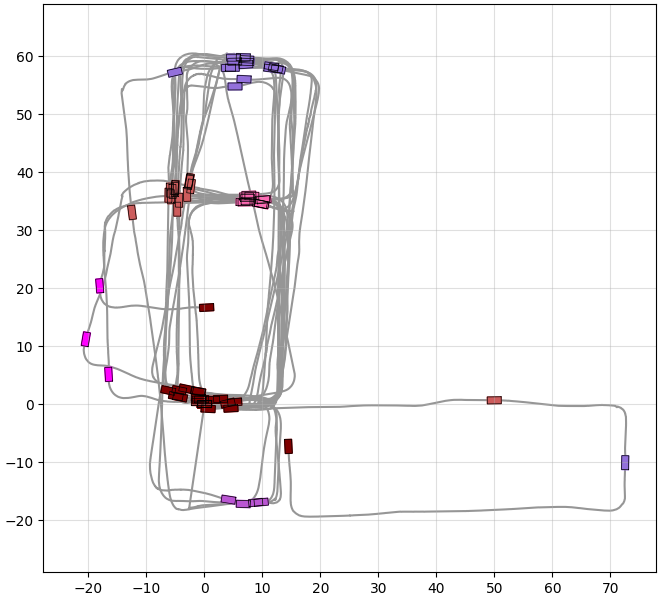}%
    \includegraphics[height=3cm]{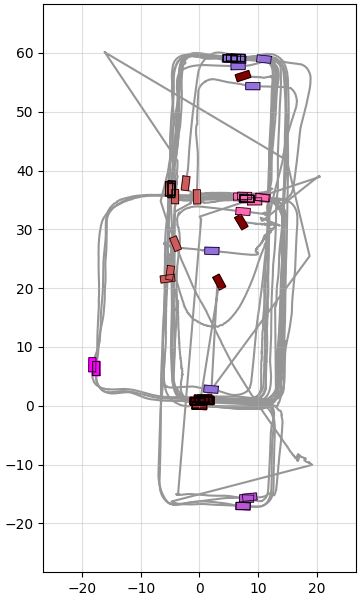}%
    \includegraphics[height=3cm]{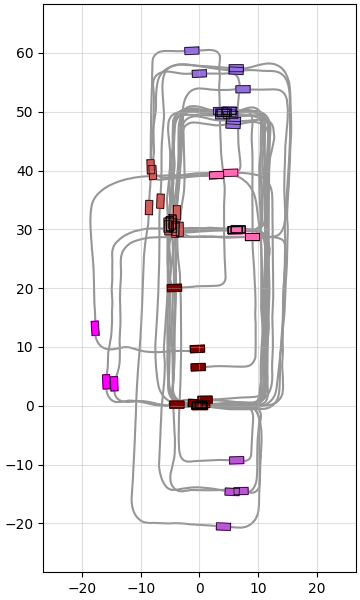}%
  }%
}
\setlength{\twosubht}{\ht\twosubbox}

% typeset

\centering

\subcaptionbox{Reference\label{fig:path_viz}}{%
  \includegraphics[height=\twosubht]{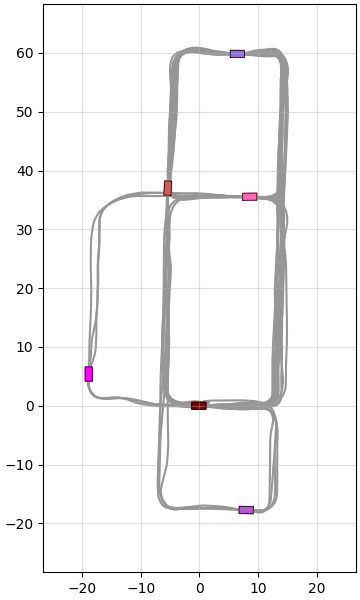}%
}
\hspace{1 em}
\rulesep
% \unskip\
% \vrule height 0.4\height depth 0.4\height
\quad
\subcaptionbox{ObVi-SLAM\label{fig:obvi_overlay}}{%
  \includegraphics[height=\twosubht]{images/TopdownTrajObVi.png}%
}\quad
\subcaptionbox{ORB-SLAM3 \label{fig:orb_overlay}}{%
  \includegraphics[height=\twosubht]{images/TopdownTrajORB.png}%
}\quad
\subcaptionbox{OA-SLAM \label{fig:oa_overlay}}{%
  \includegraphics[height=\twosubht]{images/TopdownTrajOA.png}
}\quad
\subcaptionbox{DROID-SLAM \label{fig:droid_overlay}}[7em]{%
\centering
  \includegraphics[height=\twosubht]{images/TopdownTrajDroid.png}%
}
\vspace{-0.5em}
\caption{16 trajectories through test environment, shown in meters, as estimated by the approaches with highlighted pink/purple waypoints. Performance of an approach is good when all estimates for a given waypoint are colocated.}
\label{fig:trajectory_plots}

\end{figure*}

\begin{figure*}[tbp]
\centering
\vspace{-0.5em}
\begin{subfigure}[t]{0.23\textwidth}
\centering
\includegraphics[width=\textwidth]{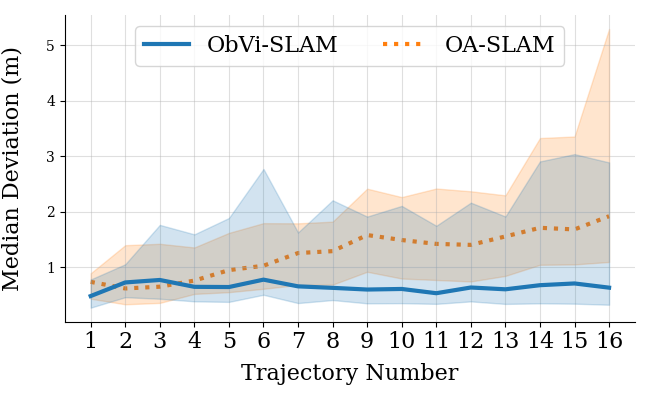}
\vspace{-1em}
\subcaption{}
\label{fig:comparison_object_pos_dev}
\end{subfigure}
\begin{subfigure}[t]{0.23\textwidth}
\centering
\includegraphics[width=\textwidth]{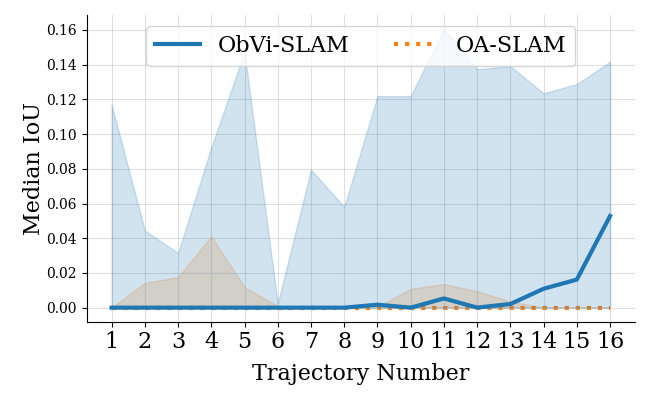}
\vspace{-1em}
\subcaption{}
\label{fig:comparison_object_iou}
\end{subfigure}
\begin{subfigure}[t]{0.23\textwidth}
\centering
\includegraphics[width=\textwidth]{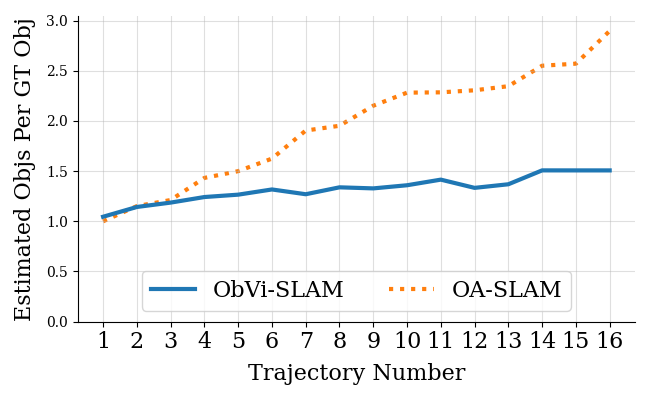}
\vspace{-1em}
\subcaption{}
\label{fig:comparison_object_false_pos}
\end{subfigure}
\begin{subfigure}[t]{0.23\textwidth}
\centering
\includegraphics[width=\textwidth]{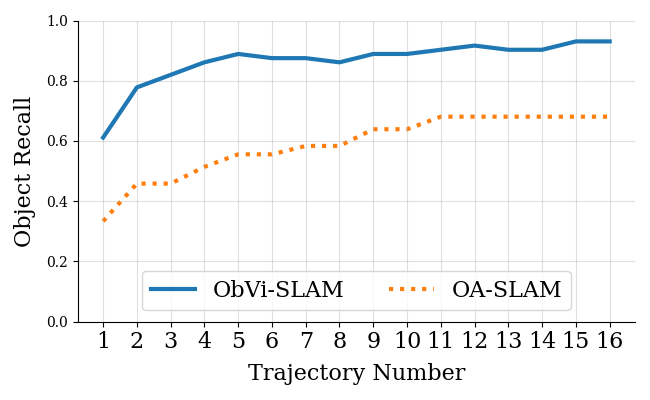}
\vspace{-1em}
\subcaption{}
\label{fig:comparison_object_false_neg}
\end{subfigure}
\caption{Object metrics for ObVi-SLAM and OA-SLAM. (\protect\subref{fig:comparison_object_pos_dev}) Lower quartile, median, and upper quartile object center error.  (\protect\subref{fig:comparison_object_iou})  Lower quartile, median, and upper quartile IoU between estimates and ground truth. (\protect\subref{fig:comparison_object_false_pos}) Estimated objects per ground truth object. (\protect\subref{fig:comparison_object_false_neg}) Recall.}
\label{fig:comparison_obj_instantiation_metrics}
\vspace{-1em}
\end{figure*}

% To understand the accuracy of individual trajectory estimates, 
To understand individual trajectory accuracy,
we computed the translational and rotational absolute trajectory error (ATE) for each approach. We first aligned each trajectory to the respective reference trajectory\hide{ before computing the error} to mitigate against early estimation error skewing results. These results are shown \hide{for all four approaches }in Fig. \ref{fig:comps_ates}. ObVi-SLAM has \hide{generally }competitive or lower rotational error\hide{ as} compared to the other approaches. \hide{There are cases where comparision approaches have lower error}Comparison approaches sometimes have lower error, but ObVi-SLAM is the only approach \hide{to have}with less than 8 degrees of average rotational error for all trajectories.  For \hide{the }translational error, ObVi-SLAM  outperforms comparision approaches on all trajectories.

\subsection{Localization Consistency Under Varying Conditions}

Long-term SLAM approaches must be able to generate estimates for key locations that are consistent over time\hide{, in spite of geometric, appearance, and viewpoint changes}. \hide{For this reason, we }We thus
assess long-term robustness by measuring consistency of  estimates for the six waypoints over the 16 trajectories 
% To assess the consistency of position estimates, we determined the centroid of all estimates for each waypoint across all trajectories and measure the distance between each estimate and the respective centroid. Likewise, we evaluated orientation consistency by calculating the mean orientation estimate for each waypoint and measuring the deviation of each estimate from the mean orientation.\AAA{cut this down}
% To assess the consistency of estimates, we measured
by measuring the deviation for each\hide{ waypoint}
estimate from the average estimate for the respective waypoint. 

Fig. \ref{fig:comps_consist_cdfs} shows the cumulative distribution functions (CDFs) for the position and orientation estimate consistency. For rotation estimate consistency, ObVi-SLAM and DROID-SLAM \cite{teed2021droid} have notably better performance than ORB-SLAM3 \cite{ORBSLAM3_TRO} and OA-SLAM \cite{zins2022oa}, with DROID-SLAM  slightly outperforming ObVi-SLAM. However, ObVi-SLAM notably outperforms all comparison approaches on position estimate consistency, having no waypoint position estimates more than 4 meters from the respective average waypoint position. 

The trajectories and waypoints estimated by LeGO-LOAM \cite{legoloam2018}  (reference), ObVi-SLAM, ORB-SLAM3 \cite{ORBSLAM3_TRO}, OA-SLAM \cite{zins2022oa}, and DROID-SLAM \cite{teed2021droid} are shown in Fig. \ref{fig:trajectory_plots}. ORB-SLAM3 lost localization in one trajectory\hide{, resulting in misalignment with the overall path,} and the remaining trajectories have \hide{varying }rotational drift,\hide{arising from feature-based maps} caused by features not being recognizable under viewpoint and lighting changes. OA-SLAM suffered relocalization failures that present as jumps in the trajectories, resulting from noisy object estimation, and DROID-SLAM has good orientation consistency, but suffers from scale inconsistency, despite use of stereo data. ObVi-SLAM, by contrast, has trajectories that match the reference well and waypoints are closely colocated.

\subsection{Object Accuracy}

To understand ObVi-SLAM's object estimation accuracy, we assess four different metrics using the map result at the end of each evaluated trajectory: 
\begin{enumerate*}
    \item position accuracy: measures the distance between estimate and ground truth object centers; 
    \item volume intersection over union (IoU): evaluates volumetric accuracy by computing the intersection of estimated and ground truth objects, divided by their union --  an IoU of 1.0 indicates a perfect estimate; 
    % \item estimated objects per ground truth: number of estimated objects generated for every detected ground truth estimate. This provides insight into the generation of duplicate and spurious objects.
    \item estimated objects per ground truth: ratio of number of estimated objects to ground truth objects, providing insight into the generation of duplicate and spurious objects; and
    \item recall: measures how many ground truth objects were \hide{successfully }estimated.
\end{enumerate*}
Ground truth data was generated by manually annotating 3D bounding boxes for 72\hide{ labeled} objects \hide{with}in a lidar-SLAM\hide{-generated} point cloud. We aligned estimates with ground truth frames and matched objects by finding the closest\hide{ corresponding} ground truth object with the same semantic class, allowing for multiple matches per ground truth object. \hide{The combined use of p}Position deviation and volume IoU together assess geometric accuracy, while the remaining metrics examine object 
instantiation false positives and \hide{false }negatives.

Fig. \ref{fig:comparison_obj_instantiation_metrics} shows these metrics for  ObVi-SLAM and OA-SLAM \cite{zins2022oa}, as the other approaches do not \hide{generate object estimates}estimate objects. 
OA-SLAM initially exhibits competitive median position deviation compared to ObVi-SLAM. However, as additional trajectories are considered, OA-SLAM's positional accuracy degrades\hide{ over time}. ObVi-SLAM maintains position estimate accuracy over time and exhibits notably better IoU for object estimates. This indicates that OA-SLAM's object estimates are less volumetrically accurate\hide{, despite closer center positions}. 
% For the initial trajectories, OA-SLAM has some lower median position deviation than ObVi-SLAM, but \hide{ObVi-SLAM's position estimate accuracy is stable with additional trajectories, whereas }OA-SLAM degrades with additional trajectories, producing less positionally-accurate objects over time. Further, ObVi-SLAM has notably better IoU for object estimates compared to OA-SLAM\hide{. This indicates}, indicating that OA-SLAM's object estimates are less volumetrically accurate, even when the center positions are closer than ObVi-SLAM's\hide{ estimates}. 
It should be noted most estimated objects are tall and thin, therefore a small amount of position error can lead to no overlap between estimates and ground truth, explaining the relatively low IoU for both approaches. In addition, ObVi-SLAM has better rates of object estimation than OA-SLAM, producing less duplicate estimates\hide{, as indicated in Fig. \ref{fig:comparison_object_false_pos},} and generating estimates for more \hide{of the }objects than OA-SLAM. The higher object recall also explains in part the initial higher upper quartile position error: ObVi-SLAM is more aggressive in generating estimates for objects, initially leading to some less accurate estimates before sufficient data 
is obtained, but resulting in fewer missed objects overall.

\subsection{Map Space Efficiency}

\begin{table}[tbp]
\centering
% \small
\caption{Size of compressed maps after specified trajectories.}
\begin{tabular}{|l|r|r|r|}
\hline
          & Trajectory 1        & Trajectory  7        & Trajectory 16        \\ \hline
ObVi-SLAM & \textbf{30 KB} & \textbf{51 KB} & \textbf{51 KB} \\ \hline
ORB-SLAM3 & 165 MB        & 1.3 GB         & -              \\ \hline
OA-SLAM   & 1.4 MB        & 8.6 MB         & 17 MB          \\ \hline
\end{tabular}

\label{table:map_space}
\end{table}

To assess the space efficiency of the maps, we evaluated the map files' compressed size after trajectories 1, 7, and 16\hide{ in our 16 trajectory sequence}\hide{. These results are}, shown in Table \ref{table:map_space}. As noted, DROID-SLAM \cite{teed2021droid} does not provide functionality for map saving\hide{saving and loading maps}, so it is excluded\hide{ from these results}. In addition, as ORB-SLAM3\cite{ORBSLAM3_TRO} could not run more than seven trajectories sequentially, the map after\hide{the seventh} trajectory 7 is the last attainable\hide{ map}\hide{ for that approach}\hide{ and thus there is no size estimate for the map after the 16th trajectory}.  From these map\hide{ file} sizes, it is clear that ObVi-SLAM's object-only map affords a much more compact and scalable representation, \hide{resulting in}with orders of magnitude smaller maps than ORB-SLAM3 and OA-SLAM \cite{zins2022oa}.

\subsection{Ablations}

\begin{figure}[tbp]
\centering
\begin{subfigure}[b]{0.49\columnwidth}
\centering
\includegraphics[width=\textwidth]{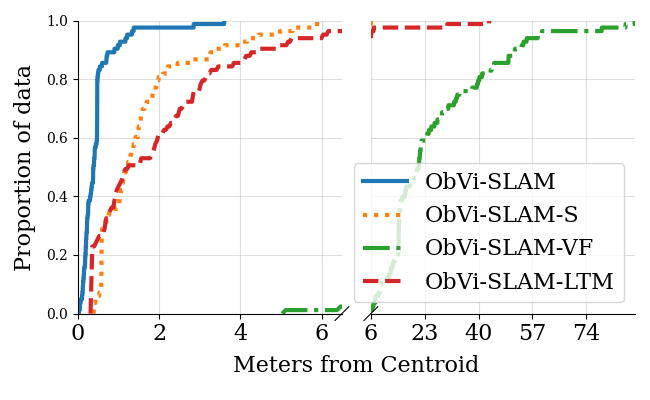}
\subcaption{CDF of position\hide{ estimate} deviation.}
\label{fig:ablation_transl_est_deviation_cdf}
\end{subfigure}
\begin{subfigure}[b]{0.49\columnwidth}
\centering
\includegraphics[width=\textwidth]{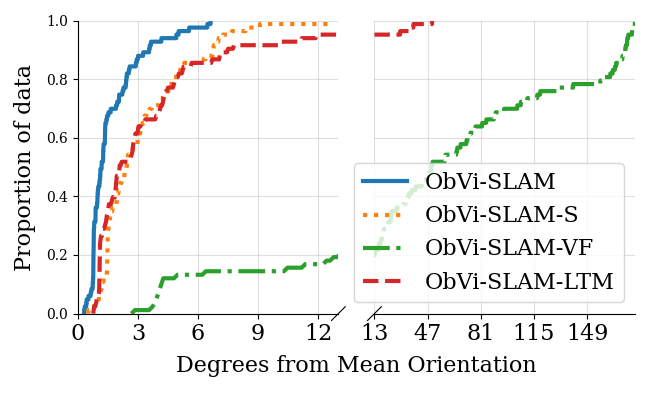}
\subcaption{CDF of orientation\hide{ estimate} deviation.}
\label{fig:ablation_orient_est_deviation_cdf}
\end{subfigure}
\caption{Position and orientation consistency for ObVi-SLAM and ablations. An optimal algorithm would quickly rise to 1.}
\label{fig:ablation_cdfs}
\vspace{-1em}
\end{figure}

In addition to comparing against other approaches, we\hide{ also} assess the performance of ObVi-SLAM when individual factors are removed\hide{ from the formulation} to understand their importance\hide{ in the proposed approach}. These are abbreviated as follows: ObVi-SLAM-S has the shape prior \hide{factor }removed; ObVi-SLAM-VF does not include the visual feature factors\hide{ in the optimization}; and ObVi-SLAM-LTM removes the long-term map\hide{ factor}, \hide{turning the problem into}resulting in a sequence of individually-evaluated trajectories.
We did not assess ObVi-SLAM without objects\hide{ factors}, as this is\hide{ functionally} equivalent to a  traditional visual SLAM approach.\hide{ORB-SLAM3 without a saved map. 
\tc{What do you mean by ``without a map"? ORB-SLAM3 has a long-term map consisted of visual features but ObVi-SLAM without object factors doesn't?}} 

The localization consistency results for ObVi-SLAM compared to the ablations is shown in Fig. \ref{fig:ablation_cdfs}. ObVi-SLAM-VF is significantly worse than all other variants, indicating the importance of visual features\hide{ estimation for trajectory accuracy}, with full ObVi-SLAM\hide{ generally slightly} outperforming the other variants. We provide ATE and object metrics for the ablated versions in our github repository\cref{footnote:website}.

\section{Conclusion and Future Work}
This paper introduces ObVi-SLAM, a long-term SLAM approach that combines the benefits of \hide{low-level }visual features and object detections to achieve scalable, long-term consistent localization amid lighting and viewpoint changes. We demonstrate that ObVi-SLAM achieves superior performance in single-trajectory and multi-trajectory localization compared to existing SLAM approaches on 16 trajectories collected over two months with significant lighting variations. 

There are \hide{a number of avenues for future work}several future directions that would build upon developments introduced in ObVi-SLAM. 
Creation of a learning-based long-term appearance descriptor for objects in the scene could improve robustness of inter-session object association. Further, integrating an object-based change detection method could aid in removing any stale data from objects incorrectly assumed to be static.
In addition, more complex factor toplogies for the long-term map prior could be explored to better capture object estimate\hide{ uncertainties and} correlations. Finally, ObVi-SLAM could be extended to allow loop closure or initial localization based on the long-term object map. 

% Future work
% \begin{itemize}
%     \item More complex forms of long-term map prior
%     \item Learned long-term object appearance descriptors for improved inter-session object association
%     \item Integration with probabilistic object maps to leverage movable objects as well
%     \item Incorporation of data-driven covariance estimates for bounding box detections
%     \item Object-based loop closure/relocalization
%     \item Change detection
% \end{itemize}

\bibliographystyle{IEEEtran}
\bibliography{bibliography}

\pagebreak

\appendices

\section{Implementation}
There are three notable aspects of implementation not detailed in the paper body:  visual feature pre-processing, two-phase optimization used for outlier rejection, and modifications for sliding window observation stability.  
\subsection{Visual Feature Pre-Processing}
% \tc{Unsure if I should motive it first}

Distant and improperly matched features can cause problems for the optimization. When detected features are far away and viewed from limited perspectives, their depth is inadequately constrained due to insufficient parallax and noise in the feature detection can cause the optimal feature estimate to be very far away. More observations should fix this problem, but if the feature is initially estimated to be extremely far away, local minima in the optimization will often cause the estimate to be stuck far away, even when a closer estimate fits the new observations better. Improper matches that can occur under repetitive textures can also cause poor initial estimates. 

To mitigate against these issues, we added a visual feature preprocessing phases. For an incoming image, if robot observed a visual feature new to the factor graph, we begin to maintain a recent observation record of this feature. After accumulating an adequate number of observations of the same feature, we compute pairwise epipolar errors between the current frame and all other associated frames stored in the record. We consider the feature observations only if their epipolar errors were small when computed against the majority of frames to reduce outliers from the feature extractions. Furthermore, we impose a minimum parallax requirement on the pending visual features to guarantee that their depth estimates are well-constrained. This requirement mandates that visual features exhibit sufficient disparities in their pixel observations as well as the robot poses from which those visual features are observed. If a visual feature meets both the epipolar error and the minimum parallax criteria, we subsequently add it to the factor graph. On the other hand, if the newly-detected visual feature has already appeared in the factor graph, we repeat a similar process to calculate the epipolar errors between the current frame and previous frames where the same feature had been observed. 
% (In this case, there was no need to assess minimum parallax requirements since the factor graph already contains enough factors to prevent unreliable depth estimations.) 
If a substantial number of the computed errors are large, we discard the current feature observation as an outlier.

\subsection{Two-Phase Optimization}
One method we have for handling outliers is a two-phase optimization for each added frame. In the first round of optimization, all factors are included and the optimization is run with loose convergence parameters so that it finishes quickly. Following this optimization, we identify the reprojection and bounding box factors with the highest costs and consider them outliers. Ignoring estimate adjustments made in the first round, we then rerun the optimization without these outlier factors so that we can base the estimate on only inlier observations. We only allow bounding box and visual feature factors to be excluded in the second phase, as the outlier exclusion primarily aims to handle incorrect data association; the long-term map prior, visual odometry factor, and semantic shape prior are not considered for outlier exclusion.

\subsection{Modifications for Sliding Window Optimization Stability}
Without special handling, as the sliding window for local optimization moves forward, features and objects can become under-constrained as they leave the window, making their estimates and the estimates for robot poses from which they are observed susceptible to dramatic changes from overfitting to the remaining observations in the window. We apply a couple modifications to the optimization to handle this. First, we set the first few poses in the sliding window constant for local optimization, so observations that are included cannot drift substantially due to the fixed observing pose. Second, we exclude features and objects from the optimization that do not have a minimum number of observations present in the sliding window. This prevents the optimization from overfitting to a small number of observations.

\section{Numerical Stability}
\subsection{Ellipsoid Projection}
As noted above, if an object intersects the x or y axes of the camera, the conic projection is not a proper ellipse. This manifests as negative values under the square roots in (\ref{eqn:x_recovery}) and/or (\ref{eqn:y_recovery}). There are two possible resulting projections: an imaginary ellipse and a hyperbola. In both cases, we currently set the error to a static fixed threshold, but we believe there are possible improvements to make in this handling. The imaginary ellipse results from the camera being inside the object, and therefore is not a valid state given any bounding box observation. An inverse barrier cost term could be added to the bounding box factor cost, so that there is minimal cost when the camera is not close to being inside the object estimate, but that the cost increases rapidly as the camera approaches the object estimate surface. However,  an object estimate could be shifted in local optimization such that the camera could start inside the object in a later global optimization,  violating the boundary condition while an observing pose is outside the local window. Thus, care must be taken in inverse barrier function selection or optimization pre-processing to prevent the estimates from being stuck in this bad state. 

\begin{figure}[tbp]
\centering
\includegraphics[width=0.83\columnwidth]{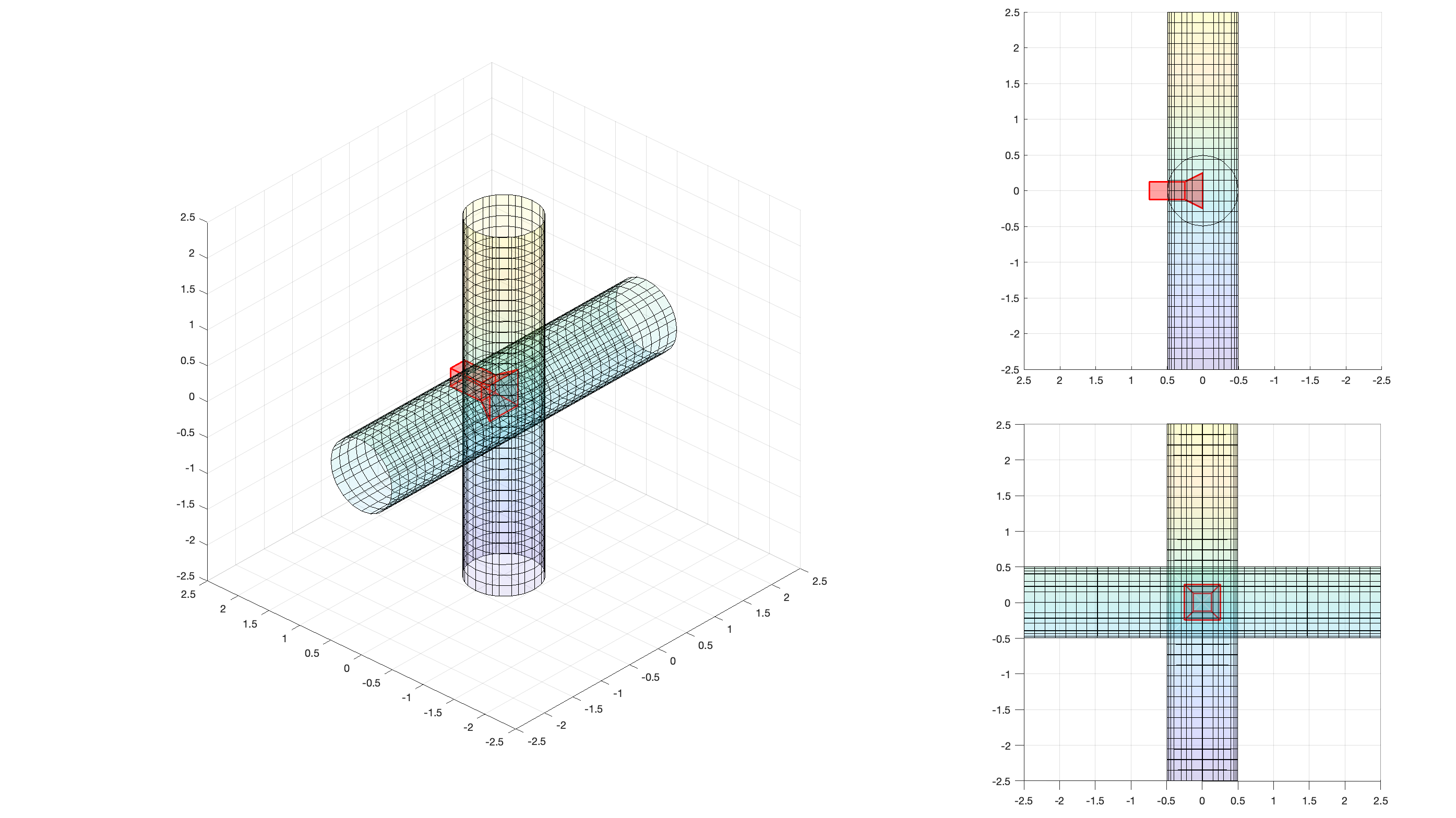}
\caption{Simplified demonstration of problematic ellipsoid projection, shown from an angle, top down, and from the front. For an object with identity orientation and height, width, and length of 1 m and camera centered at the origin, projection will fail if the object's center falls within either cylinder above. The dimensions of the object change the radii of the elliptic cylinders. }
% \AAA{Add subfigures showing top down and side views of problem state.}}
\label{fig:problem_ellipsoid}
\vspace{-1em}
\end{figure}

\begin{table*}[tbp]
\centering

\caption{Absolute Trajectory Error for Translation and Orientation. Best results are bolded and second-best are italicized.}
\begin{tabular}{|r|c|c|c|c||c|c|c|c|}
\hline
& \multicolumn{4}{|c||}{Translation (m)} &  \multicolumn{4}{c|}{Orientation (deg)}  \\ 
\hline 
  Traj Num & ObVi-SLAM & OA-SLAM & ORB-SLAM3 & DROID-SLAM & ObVi-SLAM & OA-SLAM & ORB-SLAM3 & DROID-SLAM \\ \hhline{|=|=|=|=|=||=|=|=|=|}
1 & \textbf{0.24}  & \textit{0.31} & 0.67 & 1.39 & \textit{1.7} & \textbf{1.6}  & 2.0 & 1.9 \\ \hline 

2 & \textbf{0.55}  & 3.32 & \textit{1.2} & 4.76& \textit{2.0} & \textbf{1.7}  & 2.8 & 2.4 \\ \hline 
3 & \textbf{0.84}  & 3.99 & \textit{2.36} & 4.92 & \textit{6.2} & 19.1 & 12.0 & \textbf{2.2}  \\ \hline 
4 & \textbf{0.33}  & \textit{0.6} & 1.7 & 3.96 & \textit{1.9} & \textbf{1.8}  & 2.2 & 13.6 \\ \hline 
5 & \textbf{0.33}  & \textit{1.11} & \textit{1.11} & 3.86 & 1.8 & \textbf{1.1}  & \textit{1.7} & 2.5 \\ \hline 
6 & \textbf{0.41}  & 1.31 & \textit{0.98} & 4.1 & \textit{1.9} & 2.0 & \textbf{1.7}  & 2.2 \\ \hline 
7 & \textbf{0.25}  & 1.21 & \textit{0.83} & 4.21 & \textit{1.7} & \textbf{1.3}  & \textit{1.7} & 2.2 \\ \hline 
8 & \textbf{0.47}  & 2.66 & \textit{1.8} & 3.59 & \textbf{2.2}  & 4.4 & 3.2 & \textit{2.3} \\ \hline 
9 & \textbf{0.49}  & 9.14 & 7.27 & \textit{4.79} & \textit{2.2} & 42.7 & 46.7 & \textbf{2.2}  \\ \hline 
10 & \textbf{0.58}  & \textit{0.96} & 6.82 & 3.12 & \textit{2.5} & \textbf{1.1}  & 6.7 & 3.3 \\ \hline 
11 & \textbf{0.31}  & 0.6 & \textit{0.5} & 2.05 & 2.3 & \textbf{1.6}  & \textit{1.6} & 11.1 \\ \hline 
12 & \textbf{0.36}  & 0.94 & \textit{0.67} & 7.69 & 1.9 & \textbf{1.4}  & \textit{1.7} & 2.5 \\ \hline 
13 & \textbf{0.57}  & 1.0 & \textit{0.91} & 4.84 & 7.3 & \textbf{1.0}  & \textit{1.8} & 3.1 \\ \hline 
14 & \textbf{0.87}  & 6.44 & \textit{2.29} & 6.82 & \textit{4.2} & 14.9 & 4.9 & \textbf{2.7}  \\ \hline 
15 & \textbf{0.8}  & 10.94 & \textit{1.38} & 4.55 & 4.9 & 55.3 & \textbf{2.0}  & \textit{2.3} \\ \hline 
16 & \textbf{0.18}  & 4.48 & \textit{1.35} & 3.42 & \textbf{1.6}  & 22.6 & 2.9 & \textit{2.2} \\ \hline 
Overall & \textbf{0.51}  & 4.21 & \textit{2.91} & 4.49  & \textbf{3.3}  & 18.4 & 12.3 & \textit{4.9} \\ \hline 
\end{tabular}
\label{table:full_ate}
\vspace{-1em}
\end{table*}

The hyperbolic case results from either the x or y axes of the camera intersecting the object, but not both. A simplified example is shown in Fig. \ref{fig:problem_ellipsoid}. This case can occur for long objects observed at the edge of the frame for which the object extends behind the camera. For example, we observed this when the robot passed benches. In the hyperbolic case, recovery for either the x or y limits of the projection bounding box fails and the other yields two real values, only one of which is correct; the other results from projection of the object's extent behind the camera. Because this case corresponds to valid relative object-camera estimates, a more appropriate method for handling this case would not penalize such circumstances. An improved way to include this in the cost is to identify the hyperbola branch corresponding to the portion of the object in front of the camera, and then for the unrecoverable bounds, use the intersection of that hyperbola branch with the image boundaries.

\subsection{Covariance Estimation}

As noted above, covariance $\Sigma_y$ for a least squares problem is computed from the Jacobian $J(y^*)$ with $\Sigma_y = (J^T(y^*) J(y^*))^{-1}$. The presence of an inverse in this operation means that the Jacobian must be full rank in order to compute the covariance $\Sigma_y$. Floating point calculations and approximations for matrix inverses used in optimization libraries exacerbate this problem, requiring the condition number to be below some threshold. In some cases, we observed that these requirements were not met, leading to failure to extract the covariance estimates for the ellipsoids. One such cause of these failures is objects that have the nearly same dimension in the x and y directions, leading to an under-constrained orientation. Another example is very distant features, for which changes to the distance from the camera do not notably alter the reprojected feature location.

Singularities in matrix inversion can be addressed by adding to the diagonal entries of the matrix. Using this concept to mitigate the covariance extraction problem, we added a retry system to add small residuals to the least constrained parameters if initial covariance recovery failed. When we detect such a failure, we find the degree of rank deficiency $n$ and identify the $n+1$th smallest column norm and use this as the minimum desired norm. For columns with smaller norms than the target, we add an additional prior for the individual parameter that increases the norm to match the target and then retry covariance extraction. The adjustments applied were quite small, leading to very minor differences in covariance estimation that will not meaningfully alter the optimization result in a subsequent session.

\section{Experiment Setup}
In order to assess localization consistency under varying conditions, we needed to stop at a set of waypoints consistent across the trajectories. To ensure that these locations were actually consistent, we marked the selected locations with clear tape that the wheels should be aligned to when the robot visited the waypoint. During data collection, when we stopped at a waypoint, we marked the timestamp at which the waypoint was visited. The waypoint locations were then computed using the pose for the image with the timestamp closest to the recorded timestamp.
Reference trajectories for absolute trajectory error and Fig. \ref{fig:trajectory_plots} were obtained from LeGO-LOAM \cite{legoloam2018} using a 128-beam OS1 lidar. To verify that these results were of sufficient quality to serve as ground-truth, we inspected the map consistency after each trajectory evaluation as well as the consistency of the start and end position estimates, as all trajectories started and ended at the same location. Owing to the density of the sensor and high visibility through the whole evaluation area, LeGO-LOAM was able to produce consistent maps and trajectories without observable drift with the start and end pose estimates  very well aligned. As LeGO-LOAM trajectories were run independently without a global map, the LeGO-LOAM trajectories were collectively aligned for Fig. \ref{fig:trajectory_plots} by minimizing the spread of their waypoint estimates.

\begin{figure*}[htb!]
\centering
\begin{subfigure}[t]{0.22\textwidth}
\centering
\includegraphics[width=\textwidth]{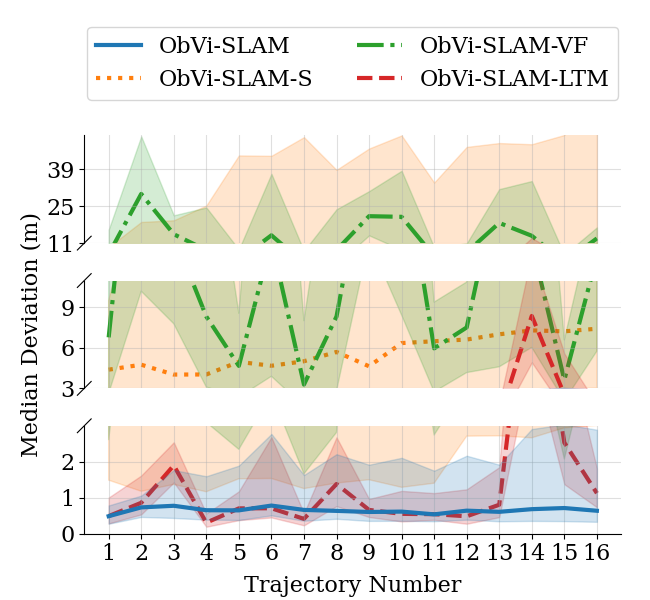}
\subcaption{}
\label{fig:ablations_object_pos_dev}
\end{subfigure}
\begin{subfigure}[t]{0.22\textwidth}
\centering
\includegraphics[width=\textwidth]{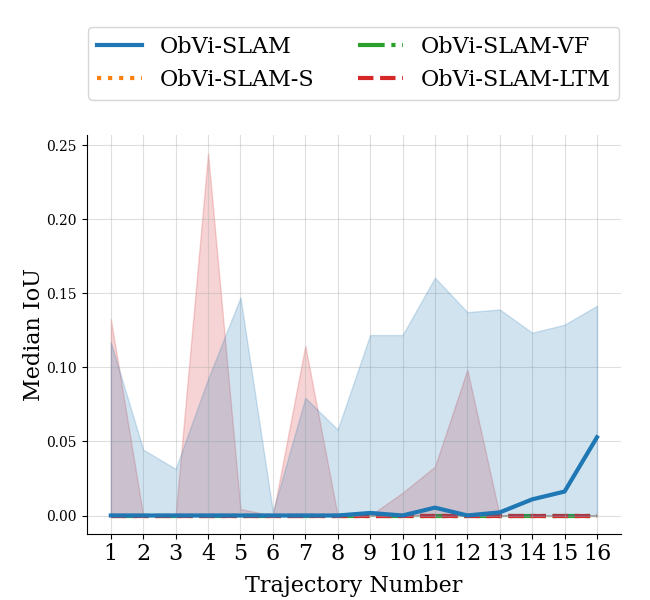}
\subcaption{}
\label{fig:ablations_object_iou}
\end{subfigure}
\begin{subfigure}[t]{0.22\textwidth}
\centering
\includegraphics[width=\textwidth]{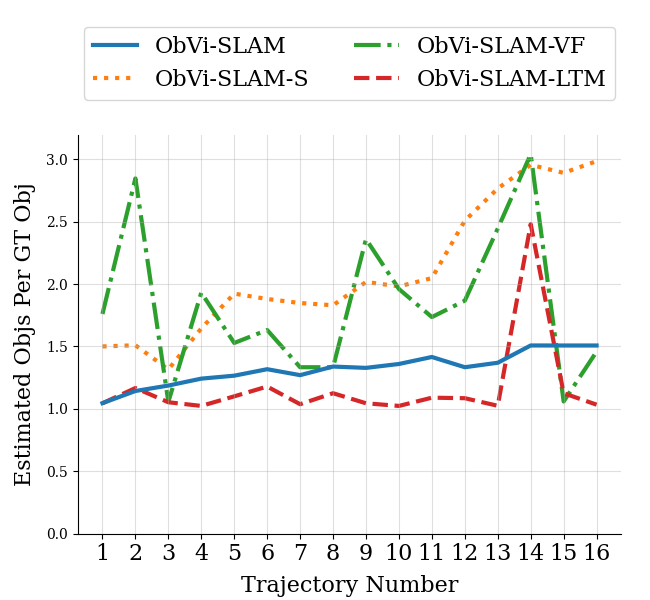}
\subcaption{}
\label{fig:ablations_object_false_pos}
\end{subfigure}
\begin{subfigure}[t]{0.22\textwidth}
\centering
\includegraphics[width=\textwidth]{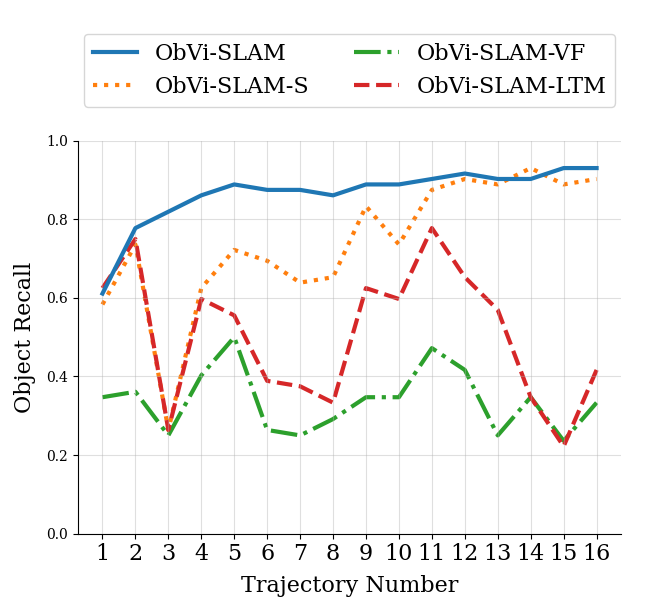}
\subcaption{}
\label{fig:ablations_object_false_neg}
\end{subfigure}
\caption{Object accuracy metrics by trajectory for ObVi-SLAM and ablations.(\protect\subref{fig:ablations_object_pos_dev}) Lower quartile, median, and upper quartile object center error.  (\protect\subref{fig:ablations_object_iou})  Lower quartile, median, and upper quartile IoU between estimates and ground truth. (\protect\subref{fig:ablations_object_false_pos}) Estimated objects per ground truth object. (\protect\subref{fig:ablations_object_false_neg}) Recall.}
\label{fig:ablations_obj_instantiation_metrics}
\end{figure*}

\begin{figure}[tb]
\centering
\begin{subfigure}[b]{0.56\columnwidth}
\centering
\includegraphics[width=\textwidth]{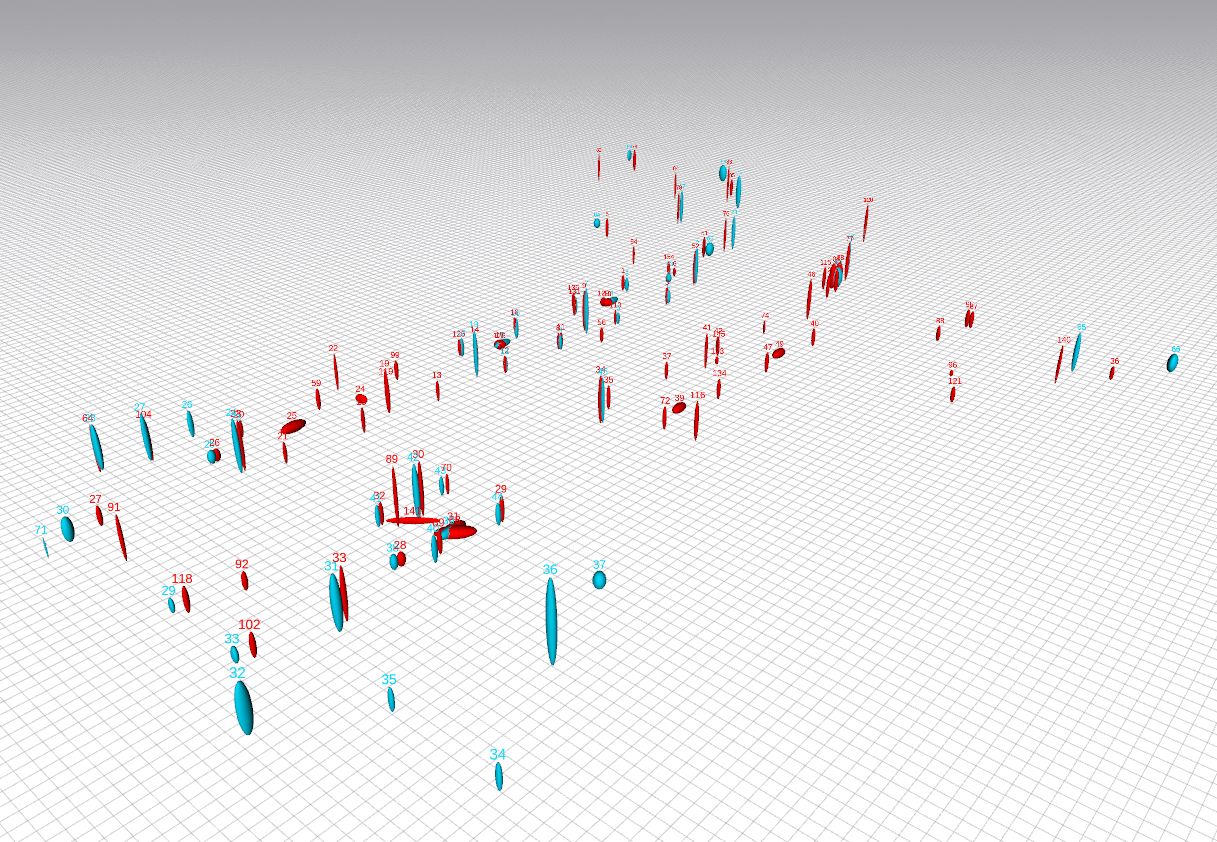}
\subcaption{ObVi-SLAM}
\label{fig:qualitative_objs_obvi}
\end{subfigure}
\begin{subfigure}[b]{0.56\columnwidth}
\centering
\includegraphics[width=\textwidth]{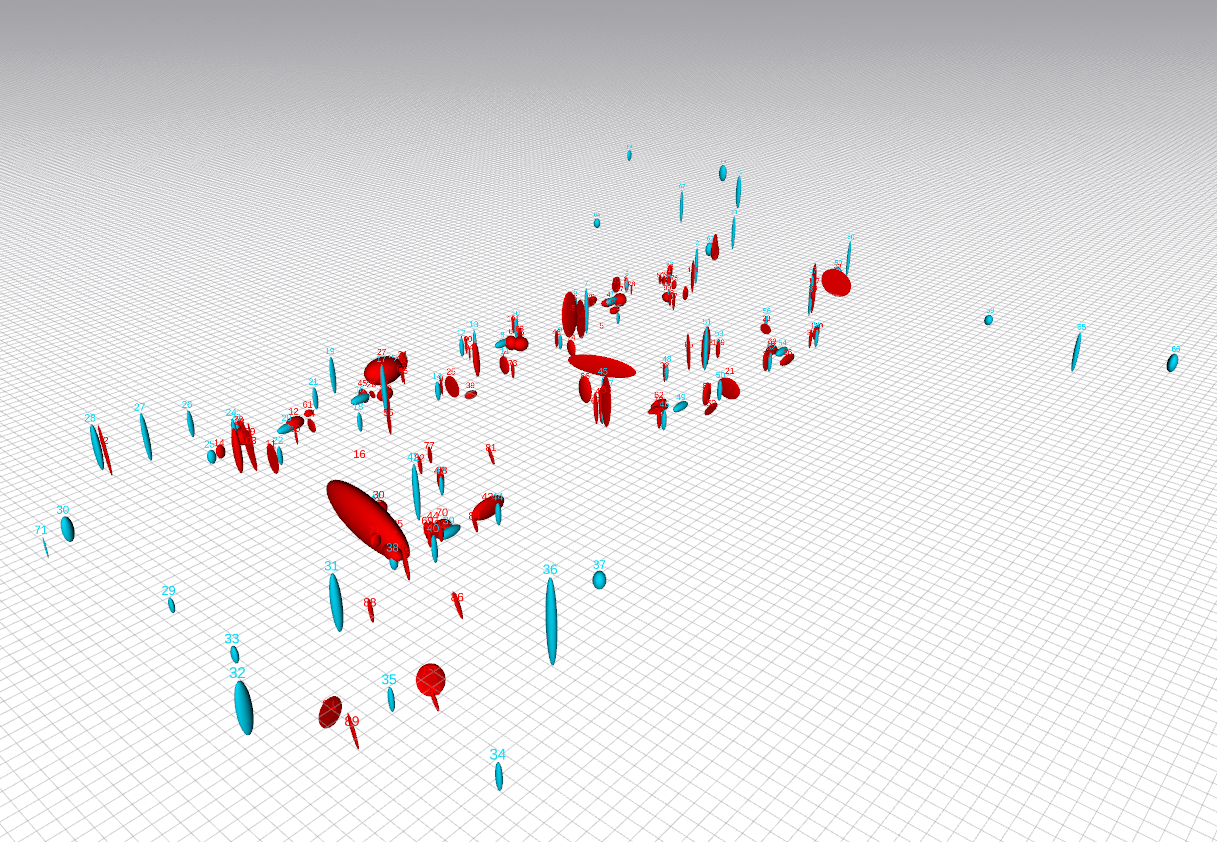}
\subcaption{OA-SLAM}
\label{fig:qualitative_objs_oa}
\end{subfigure}
\caption{Object estimates after trajectory 11. Estimated objects are shown in red and ground truth objects are in cyan.}
\label{fig:qualitative_objs}
\end{figure}

\begin{figure}[t]
\centering
\begin{subfigure}[b]{0.47\columnwidth}
\centering
\includegraphics[width=\textwidth]{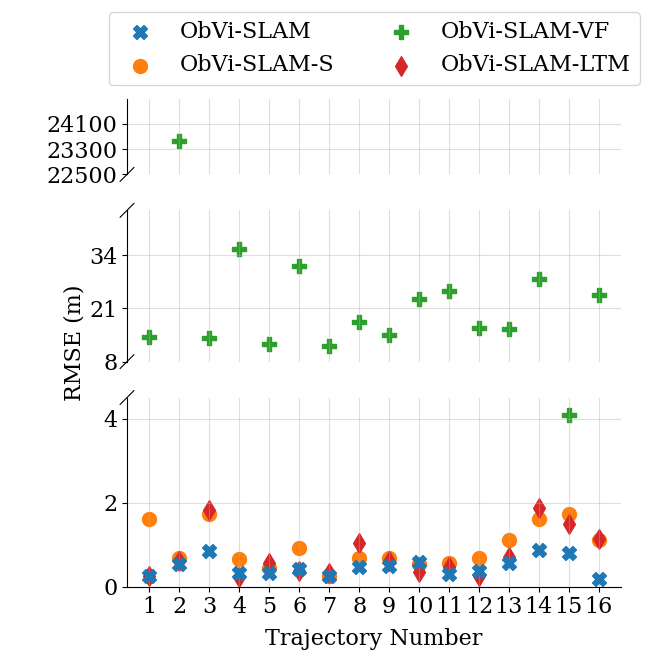}
\subcaption{Position ATE.}
\label{fig:ablation_transl_est_ate}
\end{subfigure}
\begin{subfigure}[b]{0.47\columnwidth}
\centering
\includegraphics[width=\textwidth]{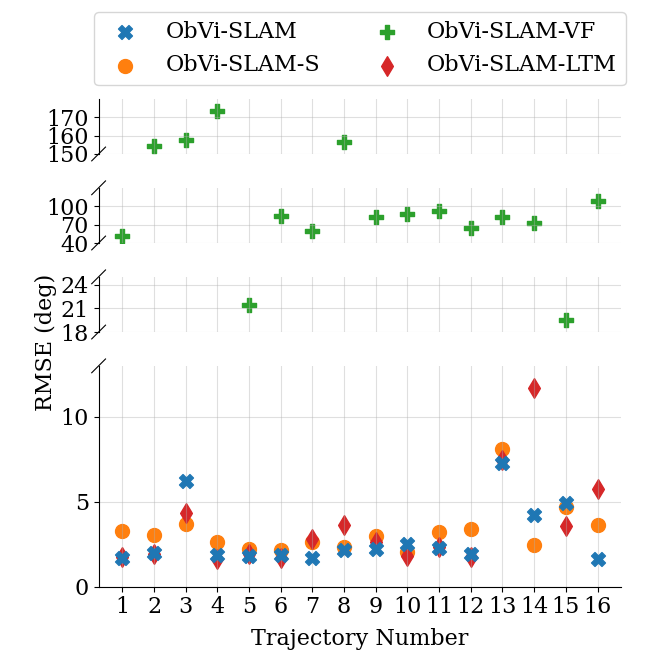}
\subcaption{Orientation ATE.}
\label{fig:ablation_orient_est_ate}
\end{subfigure}
\caption{Position and orientation ATE for ObVi-SLAM and ablations for each trajectory.}
\label{fig:ablation_ates}
\vspace{-1em}
\end{figure}

\section{Expanded Results}

Table \ref{table:full_ate} shows the absolute trajectory errors depicted in Fig. \ref{fig:comps_ates}, with the best results for each trajectory bolded. Fig. \ref{fig:qualitative_objs} shows the object estimates from ObVi-SLAM and OA-SLAM compared to the ground truth objects. These results align with the quantitative results in Fig. \ref{fig:comparison_obj_instantiation_metrics}, which show that OA-SLAM has generally worse geometrically accurate object estimates and more duplicated objects.

\section{Complete Ablation Results}

In this section, we provide the ablation study metrics not included in the main paper. Position and orientation ATE for ObVi-SLAM compared to the ablations are shown in Fig. \ref{fig:ablation_ates}. ObVi-SLAM-VF is significantly worse than all other variants, indicating the importance of visual-feature estimation for trajectory accuracy, with full ObVi-SLAM generally slightly outperforming the other variants. Object metrics for ObVi-SLAM and the ablated versions are shown in Fig. \ref{fig:ablations_obj_instantiation_metrics}. Object geometric accuracy in Figs. \ref{fig:ablations_object_pos_dev} and \ref{fig:ablations_object_iou} is poor for ObVi-SLAM-VF and ObVi-SLAM-S, demonstrating the importance of both visual features and the semantic shape prior. ObVi-SLAM-LTM occasionally demonstrates performance on par with or slightly superior to ObVi-SLAM, though it more often performs worse. This indicates that carrying object information between sessions is not always critical, but provides a floor for object geometry estimate accuracy. ObVi-SLAM-VF and ObVi-SLAM-S also perform similarly poorly on the erroneous object estimation rates shown in Figs. \ref{fig:ablations_object_false_pos} and \ref{fig:ablations_object_false_neg}. ObVi-SLAM-LTM does perform slightly better than ObVi-SLAM in the average number of estimated objects per ground-truth object, but is substantially worse at ensuring estimates exist for each ground truth object. This is not surprising, as false positives generated in one session aren't carried forward when the long-term map is removed, but also means that objects must be sufficiently viewed in every session to be included in the long-term map, which is often not the case under varying paths.

\FloatBarrier

\end{document}